\documentclass{article}

\PassOptionsToPackage{numbers, compress}{natbib}

\usepackage[boxed]{varsity}
\runningtitle{Transferability for General Reasoning: An Automated Curriculum for Multi-Domain RLVR} 

\usepackage[utf8]{inputenc} 
\usepackage[T1]{fontenc}    
\usepackage{hyperref}       
\usepackage{url}            
\usepackage{booktabs}       
\usepackage{amsfonts}       
\usepackage{nicefrac}       
\usepackage{microtype}      
\usepackage{xurl}

\usepackage{graphicx}
\usepackage{amsmath}
\usepackage{amssymb}
\usepackage{booktabs}
\usepackage{multirow}
\usepackage{makecell}
\usepackage{caption}
\usepackage{subcaption}
\makeatletter
\@namedef{ver@everyshi.sty}{}
\makeatother
\usepackage{bbm}
\usepackage{wrapfig}
\usepackage{mathtools,xparse}
\usepackage{pifont}
\usepackage{algorithmic}
\usepackage[ruled,norelsize,linesnumbered]{algorithm2e}
\usepackage{enumitem}
\usepackage{pifont}
\usepackage{array}
\usepackage{scalerel,xparse}
\usepackage{colortbl}
\usepackage{amsthm}
\usepackage[english]{babel}
\usepackage{bm}
\usepackage{mdframed}
\usepackage{tabularx} 

\usepackage[most,skins,theorems]{tcolorbox}
\tcbset{
  aibox/.style={
    width=0.95\linewidth,
    top=6pt, bottom=3pt,
    colback=blue!6!white, colframe=black, colbacktitle=black,
    enhanced, center,
    attach boxed title to top left={yshift=-0.05in,xshift=0.1in},
    boxed title style={boxrule=0pt,colframe=white,},
    width=\columnwidth,
  }
}
\tcbuselibrary{breakable}

\newcommand{\mytitle}{Transferability for General Reasoning: \\An Automated Curriculum for Multi-Domain RLVR}

\newtcolorbox[]{AIbox}[2][]{%
  aibox,
  title={#2~\thetcbcounter},
  label=#1
}

\newcommand{\alg}{\textbf{\texttt{TAC}}\xspace}
\newcommand{\fullalg}{Transfer-Aware Curriculum\xspace}
\usepackage[multiple]{footmisc}
\title{\mytitle}

\author{Yongjin Yang${}^{1}$ \quad Jiarui Liu${}^{2}$ \quad Yinghui He${}^{3}$ \quad Lechen Zhang${}^{4}$ \\ Bernhard Sch\"olkopf${}^{5,6}$ \quad Zhijing Jin${}^{1,6,7}$ \\ Jinesis Lab, University of Toronto \& Vector Institute${}^{1}$ \\ Carnegie Mellon University${}^{2}$ \quad Princeton University${}^{3}$ \\ University of Illinois Urbana-Champaign${}^{4}$ \quad ELLIS Institute T\"ubingen${}^{5}$ \\ Max Planck Institute for Intelligent Systems${}^{6}$ \quad EuroSafeAI${}^{7}$ \\
\texttt{\{yjyang,zjin\}@cs.toronto.edu}
}

\paperlinks{\href{https://github.com/YangYongJin/transfer-aware-curriculum}{\texttt{github.com/YangYongJin/transfer-aware-curriculum}}}

\begin{document}

%
\begin{abstract}
Reinforcement learning with verifiable rewards (RLVR) has been extended from single-domain training to multi-domain reasoning suites spanning mathematics, programming, and science. 
However, the training curriculum (how often each domain is sampled) is typically fixed or hand-tuned, even though reasoning skills transfer unevenly across domains. 
Existing learnability-based curricula adapt to where the policy is currently improving, but are blind to whether a gradient step on the selected domain benefits the remaining domains. 
In this paper, we propose \fullalg~(\alg), a bandit-style online curriculum that prioritizes domains whose updates broadly benefit the rest of the training suite.
\alg~repurposes signals already produced by RL training: per-domain advantages capture local learnability, and projected gradients, taken from the GRPO step being computed, estimate cross-domain transferability via gradient-geometry alignment, at negligible cost ($<$1\% wall-clock overhead).
Across a six-domain reasoning suite, \alg~achieves the best macro-averaged accuracy on both \texttt{Qwen3-1.7B} and \texttt{Llama3.2-3B}, outperforming proportional random sampling, a hand-designed schedule, and a learnability-only bandit, and improving over the last of these by up to 2.8 points (10\% relative).
Ablations show performance degrades sharply when the transferability term is removed, and \alg~remains robust on imbalanced training mixtures where learnability-only curricula over-commit to dominant domains. 
Our findings establish cross-domain transferability as a key signal for curriculum design in multi-domain RLVR.
\end{abstract}

\section{Introduction}
\label{sec:introduction}

Reinforcement learning with verifiable rewards~(RLVR) has become a central tool for improving the reasoning capabilities of large language models~(LLMs), producing substantial gains on benchmarks where correctness can be checked automatically~\citep{guo2025deepseek, shao2024deepseekmath}.
Motivated by these successes, recent work has extended RL training beyond single-domain setups toward broader multi-domain reasoning suites spanning domains such as mathematics, programming, and science~\citep{ma2025generalreasoner, chengrevisiting}.
The goal of this line of work is a single policy that reasons competently across heterogeneous tasks rather than one specialized in a narrow slice.

\begin{figure*}[t]
    \centering
    \begin{subfigure}[t]{0.33\linewidth}
        \centering
        \includegraphics[width=\linewidth]{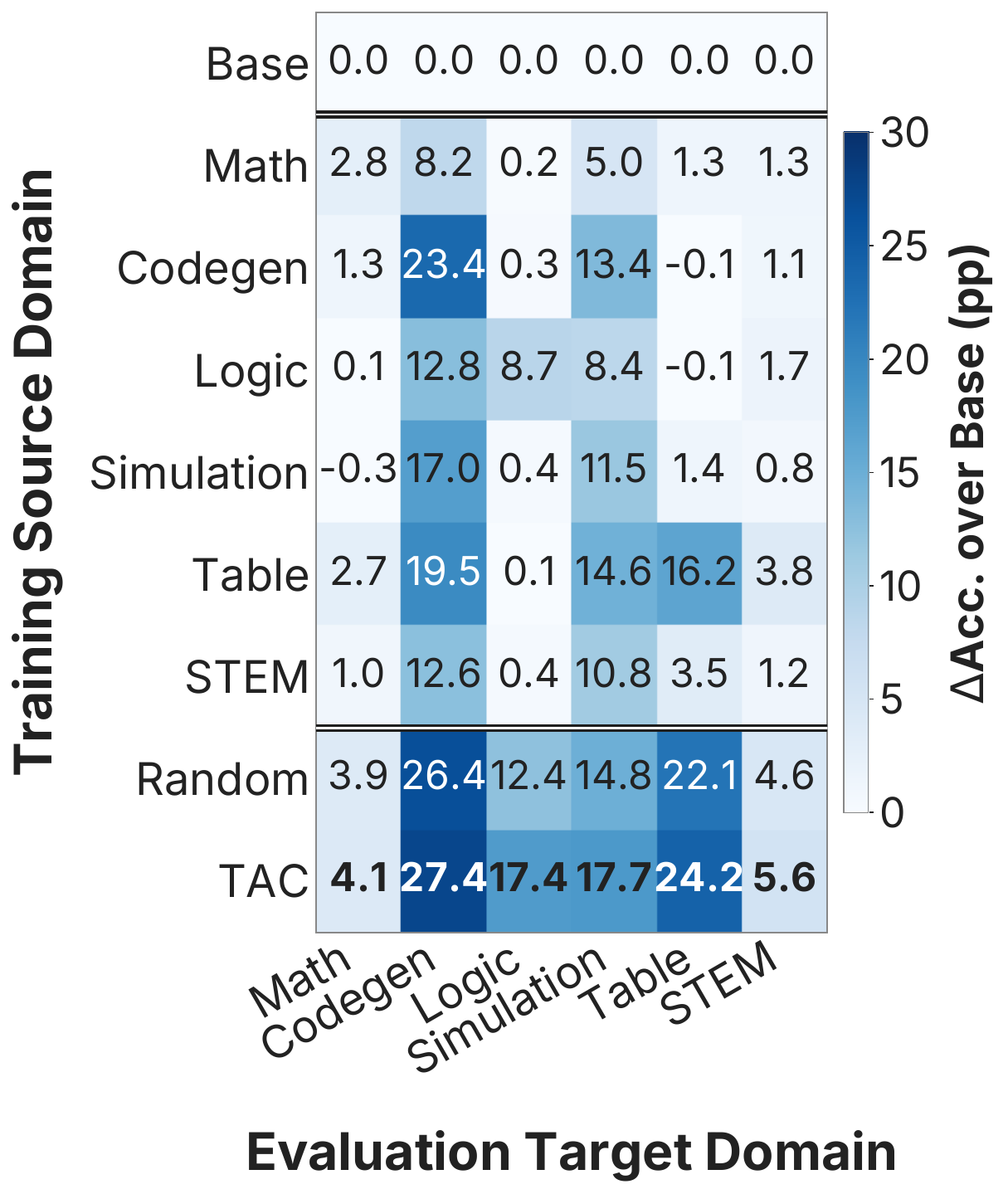}
        \caption{Cross-domain transfer matrix.}
        \label{fig:transfer_matrix}
    \end{subfigure}
    \hfill
    \begin{subfigure}[t]{0.64\linewidth}
        \centering
        \includegraphics[width=\linewidth]{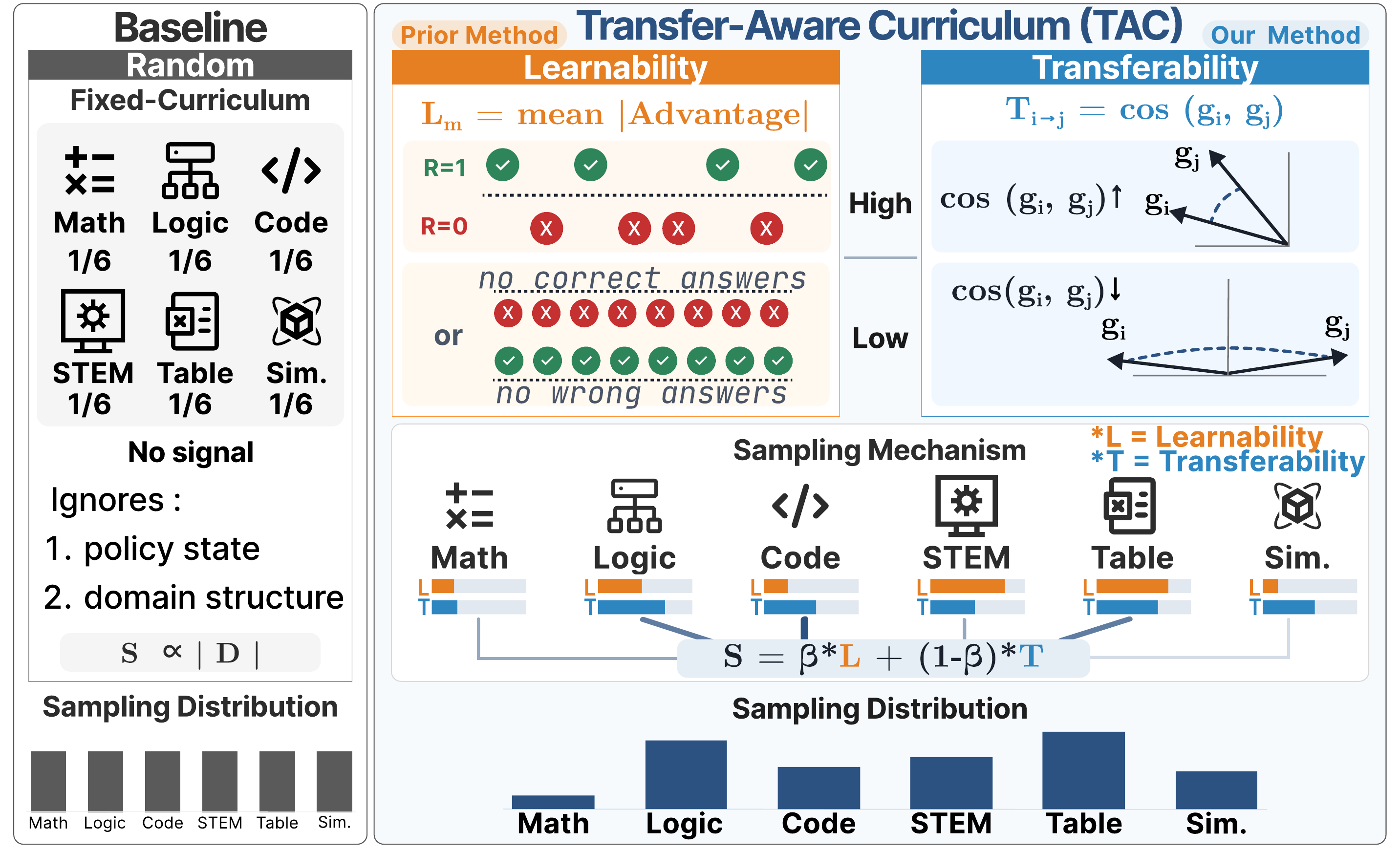}
        \caption{\alg vs. baseline and prior-work selection signals.}
        \label{fig:method_overview}
    \end{subfigure}
    \caption{\textbf{(a)} Each cell shows the accuracy gain (pp) on a target domain after RL-training on a single source domain (1{,}000 queries, two epochs). Off-diagonal transfer varies sharply by source. The bottom rows show that \alg improves over \textit{Random} across all six domains. \textbf{(b)} \alg combines a learnability signal with a gradient-cosine transferability signal that captures whether updates on a domain align with those of the other domains, mixed into per-arm scores via a single coefficient $\beta$.}
    \label{fig:overview}
    \vspace{-12pt}
\end{figure*}

Building a strong multi-domain reasoner, however, is not merely a matter of pooling data from every domain. Recent analyses show that reasoning training transfers unevenly across tasks, with effects depending strongly on the source domain~\citep{huan2025does, li2025can, chengrevisiting}. We observe the same pattern in our own setting (Figure~\ref{fig:transfer_matrix}): RL-training on different single domains yields markedly different transfer profiles across the remaining domains. For example, given the same training budget, RL on \textit{table} improves \textit{simulation} accuracy by 14.6 percentage points, while RL on \textit{math} improves it by only 5.0. Methodological responses have so far focused on optimization- and loss-level interventions: per-domain gradient alignment~\citep{liang2026boosting} or per-task loss reweighting for multi-task GRPO~\citep{ramesh2026multi}. The orthogonal question of \emph{which domain to sample at each step}, and how that choice should depend on cross-domain transferability, has received far less attention.

Existing curricula for LLM reasoners~\citep{chen2025self, wang2025dump, jiang2025vcrl} operate at the sampling level, but answer only half of this question. 
They prioritize domains whose on-policy advantages~\citep{chen2025self, wang2025dump} or reward variances indicate active learning~\citep{jiang2025vcrl}, telling us \emph{which domain the policy can presently learn from}, but not whether a gradient step on that domain also \emph{benefits the remaining training domains}. 
A highly learnable domain may be narrowly scoped, so that gains on it fail to transfer; a less obviously learnable domain may nonetheless yield updates whose direction broadly improves the policy. 
A curriculum blind to cross-domain transfer can over-commit to locally rich but globally narrow domains, leaving a large share of the achievable cross-domain improvement on the table.
\vspace{-5pt}
\paragraph{Contributions.}
In this paper, we propose \fullalg~(\alg), a curriculum for multi-domain RL that introduces \emph{cross-domain transferability} as a first-class signal for domain selection. While prior curricula prioritize domains the policy can presently learn from, \alg additionally asks whether a gradient step on a selected domain benefits the \emph{remaining} training domains, and steers sampling accordingly. 
Figure~\ref{fig:transfer_matrix} previews the result: uniform multi-domain sampling (\textit{Random}) already outperforms any single-source curriculum on every target, and \alg~widens this gap further across all six domains. The two signals come apart in practice: the most \emph{learnable} domain is often not the most \emph{transferable}. 
Notably, \textit{math}, the domain RLVR leans on most, ranks among the \emph{least} transferable in our suite, and \alg~down-weights it accordingly.
Our specific contributions are as follows:
\begin{enumerate}[leftmargin=1.2em, itemsep=0.7mm, topsep=0.7pt]
    \item \textbf{Transfer-Aware Bandit Curriculum.} We formulate multi-domain RL training as a multi-armed bandit over domains~(\S\ref{subsec:bandit}), whose per-arm feedback combines an on-policy learnability term~(\S\ref{subsec:learnability}) derived from GRPO advantages with a novel gradient-based transferability term, mixed via a single coefficient $\beta$~(\S\ref{subsec:tac_rl}). The resulting method, \alg, jointly prioritizes domains that are currently learnable \emph{and} broadly beneficial to the rest of the training set.

    \item \textbf{Self-Supervised Transferability Signal.} We introduce a gradient-geometry estimator of cross-domain transferability (\S\ref{subsec:transferability}). At each training step, we maintain a per-domain exponential moving average of projected gradients; every $K_c$ steps, the pairwise cosine similarities between these EMAs serve as the transferability signal fed back to the bandit. The signal is computed entirely from gradients already produced by RL training, requiring no held-out probes, extra rollouts, or oracle annotations, and adapts with the policy as it evolves.

    \item \textbf{Empirical Gains on Multi-Domain Reasoning.} Across a six-domain reasoning suite spanning mathematics, programming, logic, simulation, tables, and stem, \alg~consistently outperforms proportional sampling, a hand-designed math-to-others schedule, and a learnability-only bandit, improving macro-averaged accuracy across 14 evaluation benchmarks by 1.6--2.8 points (up to 10\% relative) on \texttt{Qwen3-1.7B} and \texttt{Llama3.2-3B} at $<$1\% wall-clock overhead (\S\ref{sec:experiments}, Appendix~\ref{app:complexity}). Ablations isolate the transferability term, and \alg~remains robust under data-budget skew where learnability-only curricula over-commit to dominant domains.
\end{enumerate}
\section{Preliminaries}
\label{sec:preliminary}

\subsection{Multi-Domain Reinforcement Learning for Reasoning}
\label{subsec:multitask_rl}
Let $\mathcal{D} = \{D_1, \dots, D_M\}$ be a collection of reasoning domains, where
each $D_m$ consists of prompt--answer pairs $(x, y^*)$ with verifiable targets.
Given a query $x$, the policy $\pi_\theta$ generates a response
$o \sim \pi_\theta(\cdot \mid x)$, evaluated by a sparse correctness reward
$r(o, y^*) = \mathbb{I}[\mathrm{Ans}(o)=y^*]$.
Multi-domain RL seeks a single policy maximizing expected reward under a sampling
distribution $\mu \in \Delta^{M-1}$ over domains:
\begin{equation}
  J(\theta;\,\mu)
  = \sum_{m=1}^{M} \mu_m \;
    \mathbb{E}_{(x,y^*)\sim D_m,\, o\sim\pi_\theta(\cdot\mid x)}\bigl[
      r(o, y^*)
    \bigr].
  \label{eq:multitask_obj}
\end{equation}
\subsection{Group Relative Policy Optimization (GRPO)}
\label{subsec:grpo}
We optimize $\theta$ with Group Relative Policy Optimization
(GRPO;~\citep{shao2024deepseekmath}). For a query $x$, GRPO draws $K$ rollouts
$\{o^{(k)}\}_{k=1}^{K}$ from the old policy $\pi_{\theta_{\text{old}}}$, where each $o^{(k)} = (o_1^{(k)}, \dots, o_{|o^{(k)}|}^{(k)})$ is the token sequence of a sampled response, and computes group-normalized advantages
\begin{equation}
  A^{(k)} = \frac{r^{(k)} - \bar{r}}{\sigma_r + \epsilon },
  \qquad
  \bar{r} = \tfrac{1}{K}{\textstyle\sum_k} r^{(k)},
  \quad
  \sigma_r = \sqrt{\tfrac{1}{K}{\textstyle\sum_k}(r^{(k)}-\bar{r})^2},
  \label{eq:grpo_adv}
\end{equation}
yielding the clipped surrogate objective
\begin{equation}
  \mathcal{L}_{\mathrm{GRPO}}(\theta)
  = -\mathbb{E}_x\!\left[
      \frac{1}{K}\sum_k\sum_t
      \min\!\bigl(p_t^{(k)} A^{(k)},\;
                  \mathrm{clip}(p_t^{(k)}, 1 \pm \varepsilon)\, A^{(k)}\bigr)
    \right],
  \label{eq:grpo_loss}
\end{equation}
where
$p_t^{(k)} = \pi_\theta(o_t^{(k)}\mid x,o_{<t}^{(k)})
 / \pi_{\theta_{\mathrm{old}}}(o_t^{(k)}\mid x,o_{<t}^{(k)})$ is the per-token importance ratio. We follow the DAPO recipe~\citep{yu2025dapo} and omit the KL regularizer used in the original formulation.
\subsection{Curriculum as a Multi-Armed Bandit}
\label{subsec:bandit}
A fixed sampling mixture $\mu$ is generally suboptimal: under a given policy, the optimization signal each domain provides, and how well its updates transfer to the others, drifts as training proceeds. We therefore adapt $\mu^{(t)}$ online from the training history $\mathcal{H}_{t-1}$.
\vspace{-5pt}
\paragraph{Bandit formulation.}
We cast domain selection as a multi-armed bandit in which each domain $D_m$ corresponds to an arm. At each step $t$ the curriculum (i) samples an arm $m_t$ from a distribution over per-arm value estimates $\{Q_m^{(t-1)}\}$; (ii) draws a single-domain minibatch $\mathcal{B}_t \subset D_{m_t}$ and applies one GRPO update; (iii) observes a scalar feedback signal $S_{m_t}^{(t)}$; and (iv) refreshes the pulled arm's estimate via an exponential moving average,
\begin{equation}
  Q_{m_t}^{(t)} = (1-\alpha)\,Q_{m_t}^{(t-1)} + \alpha\,S_{m_t}^{(t)},
  \label{eq:q_update}
\end{equation}
with bandit learning rate $\alpha\in(0,1]$. By default the pulled arm is refreshed and the others retain their previous estimates; \alg extends step~(iv) to also refresh unsampled arms whose transferability has been recomputed, while leaving the EMA form of Eq.~(\ref{eq:q_update}) unchanged (\S\ref{subsec:tac_rl}). The EMA absorbs the non-stationarity that arises as $\pi_\theta$ evolves.
\vspace{-5pt}
\paragraph{Arm selection.}
We sample arms from a Boltzmann distribution over UCB-augmented values:
\begin{equation}
  \mu_m^{(t)} \;\propto\; \exp\!\left(\frac{1}{\tau}\!\left[\,Q_m^{(t-1)} + \frac{c}{\sqrt{n_m^{(t-1)} + 1}}\,\right]\right),
  \label{eq:sampling}
\end{equation}
where $n_m^{(t-1)} = \sum_{s<t}\mathbb{I}[m_s=m]$ is the visit count, $c>0$ scales the exploration bonus following the UCB form~\citep{auer2002using}, and $\tau$ is a softmax temperature. We use a deliberately soft temperature $\tau=0.85$, which keeps the UCB exploration bonus influential relative to the small Q-value gaps, so domains whose value dipped early are revisited rather than starved.
\vspace{-5pt}
\paragraph{Design degree of freedom.}
Under this template, the curriculum is fully specified by the per-arm feedback signal $S_m^{(t)}$: the bandit machinery in Eqs.~(\ref{eq:q_update})--(\ref{eq:sampling}) is invariant to its choice, while $S_m^{(t)}$ determines what the curriculum optimizes for. Prior bandit curricula instantiate $S_m^{(t)}$ as a \emph{learnability} term~\citep{chen2025self}, prioritizing domains the policy is presently improving on but blind to whether updates on the chosen domain benefit the others. In Section~\ref{sec:method} we keep the same template and introduce a feedback signal that pairs learnability with a gradient-based measure of \emph{cross-domain transferability}.
\section{Method}
\label{sec:method}

Selecting which domain to train on at each step of multi-domain RL hinges on two questions. 
First, how much \emph{optimization signal} does the domain currently provide? 
Second, and central to our work, how well does the resulting gradient step \emph{transfer} to the remaining domains? A curriculum tracking only the first over-commits to locally rich but globally narrow domains; one tracking only the second starves the policy of usable signal. 

We propose \fullalg~(\alg), a curriculum addressing both within the bandit framework of Section~\ref{subsec:bandit}. \alg specifies $S_m^{(t)}$ as a weighted combination of \textit{(1)} a \textbf{local learnability term}~(\S\ref{subsec:learnability}) derived from on-policy GRPO advantages, and \textit{(2)} a \textbf{global transferability term}~(\S\ref{subsec:transferability}) estimating how well an update on $D_m$ aligns with updates on the others. The full procedure is summarized in Figure~\ref{fig:method_overview} and Algorithm~\ref{alg:tac}; details follow in Section~\ref{subsec:tac_rl}.

\subsection{Local Learnability Signal}
\label{subsec:learnability}

For a minibatch $\mathcal{B}_t\subset D_{m_t}$ of size $B$, the mean absolute
GRPO advantage
\begin{equation}
  L_{m_t}^{(t)} \;=\; \frac{1}{BK}\sum_{b=1}^{B}\sum_{k=1}^{K}\bigl|A_b^{(k)}\bigr|
  \label{eq:learnability}
\end{equation}
serves as an on-policy proxy for learnability~\citep{chen2025self, wang2025dump}: with binary rewards, it is large when the rollout group contains a mix of successes and failures, the regime of active improvement, and collapses to zero when all $K$ rollouts share the same reward, either because $D_m$ is saturated or because it is presently beyond the policy's capacity. Because its raw scale varies across domains and training
stages, we feed a running z-score into the curriculum,
\begin{equation}
  \hat{L}_{m_t}^{(t)}
  \;=\; \frac{L_{m_t}^{(t)} - \mu_L^{(t)}}{\sigma_L^{(t)} + \epsilon},
  \label{eq:learnability_norm}
\end{equation}
where $\mu_L^{(t)}$ and $\sigma_L^{(t)}$ are EMAs of the observed $L$ values' mean and standard deviation; during the first few observations, before these statistics are reliable, we feed the raw $L_{m_t}^{(t)}$ to the curriculum instead of the z-score (Appendix~\ref{app:im_details}).

\subsection{Gradient-Based Transferability}
\label{subsec:transferability}
A good curriculum should not only select domains on which the policy is currently learning, but also favor domains whose gradient updates are directionally aligned with those of the other training domains. We estimate this alignment directly from training gradients. 
\vspace{-5pt}
\paragraph{Projected-gradient representation.}
Full gradient vectors are high-dimensional and expensive to compare across domains. Following~\citet{panigrahi2026in}, we sketch gradients into a shared low-dimensional space via TRAK-style random projections~\citep{park2023trak}. Let $P\in\mathbb{R}^{d\times r}$ be the fixed projection matrix with $r\ll d$, where $d$ is the dimensionality of a designated parameter subset (the last $N$ transformer layers). Given the GRPO gradient $g_t = \nabla_\theta\mathcal{L}_{\mathrm{GRPO}}(\theta^{(t)};\mathcal{B}_t)$ restricted to that subset, the projected representation is the unit-normalized sketch
\begin{equation}
  \mathbf{v}_t \;=\; \frac{P^\top g_t}{\bigl\|P^\top g_t\bigr\|_2} \;\in\;\mathbb{R}^r.
  \label{eq:proj_grad}
\end{equation}
The $\ell_2$ normalization makes the per-domain state below a pure average of update \emph{directions}: without it, a single step with an unusually large gradient dominates the accumulated state for many steps, and the cosine comparisons then mostly reflect that one outlier. We deliberately apply no response-length rescaling to $\mathbf{v}_t$: response length varies systematically across domains, so magnitude corrections based on it inject a persistent domain-level bias into the accumulated direction rather than removing noise. The projection matrix $P$ is generated once with a fixed seed and reused throughout training; in practice we project the gradient of the final PPO mini-batch of each step, and the smoothing below absorbs the additional variance (Appendix~\ref{app:im_details}).

\vspace{-5pt}
\paragraph{Per-domain gradient state.}
For each domain $D_m$ we maintain an exponential moving average of its projected gradient, updated only when $D_m$ is selected:
\begin{equation}
  \mathbf{h}_{m_t}^{(t)}
  \;=\; \gamma\,\mathbf{h}_{m_t}^{(t-1)} + (1-\gamma)\,\mathbf{v}_t,
  \label{eq:grad_ema}
\end{equation}
with decay $\gamma\in[0,1)$; the first observation for a domain initializes its state directly, and unselected domains retain their previous EMAs. Each $\mathbf{h}_m^{(t)}$ thus accumulates a smoothed representation of the gradient directions most recently induced by $D_m$.
\vspace{-5pt}
\paragraph{Pairwise transferability.}
A domain scores highly if its gradient state is aligned with those of the other active training domains. Every $K_c$ steps, we recompute the raw pairwise transfer of \emph{every} domain with an initialized gradient state as the mean cosine similarity against all others,
\begin{equation}
  \rho_m^{(t)}
  \;=\; \frac{1}{|\mathcal{A}_t|-1}\sum_{\substack{j\in\mathcal{A}_t\\ j\neq m}}
    \frac{\langle\mathbf{h}_m^{(t)},\,\mathbf{h}_j^{(t)}\rangle}
         {\|\mathbf{h}_m^{(t)}\|_2\,\|\mathbf{h}_j^{(t)}\|_2 + \epsilon},
  \label{eq:raw_transfer}
\end{equation}
where $\mathcal{A}_t$ is the set of domains with initialized gradient states. Note that $\rho_m^{(t)}$ changes at every comparison even for domains that were not sampled in between, because the \emph{other} domains' gradient states have moved.
\vspace{-5pt}
\paragraph{Temporal smoothing.}
To damp this step-level variance, each domain's raw cosine is first passed through a per-domain EMA with smoothing coefficient $\delta\in[0,1)$, refreshed for all $m\in\mathcal{A}_t$ at comparison steps:
\begin{equation}
  \tilde{\rho}_m^{(t)}
  \;=\;
  \begin{cases}
    \delta\,\tilde{\rho}_m^{(t-1)} + (1-\delta)\,\rho_m^{(t)},
      & m \in \mathcal{A}_t \text{ at refresh steps } (t \bmod K_c = 0),\\[2pt]
    \tilde{\rho}_m^{(t-1)},
      & \text{otherwise;}
  \end{cases}
  \label{eq:transfer_ema}
\end{equation}
between comparisons the smoothed estimate carries the signal forward.
\vspace{-5pt}
\paragraph{Cross-domain normalization.}
What matters for domain selection is not a domain's absolute cosine, which is uniformly small in the projected subspace (Appendix~\ref{app:pairwise}), but how it compares to the \emph{other} domains right now. We therefore map the smoothed cosines to a bounded relative score via min--max normalization with a floored, EMA-smoothed scale:
\begin{equation}
  s^{(t)} \;=\; \delta_s\,s^{(t-1)} + (1-\delta_s)\Bigl(\max_{j\in\mathcal{A}_t}\tilde{\rho}_j^{(t)} - \min_{j\in\mathcal{A}_t}\tilde{\rho}_j^{(t)}\Bigr),
  \qquad
  T_m^{(t)}
  \;=\; \mathrm{clip}\!\left(
    \frac{\tilde{\rho}_m^{(t)} - \min_{j\in\mathcal{A}_t}\tilde{\rho}_j^{(t)}}
         {\max\bigl(s^{(t)},\,s_{\min}\bigr)},
    \,0,\,1\right),
  \label{eq:minmax_transfer}
\end{equation}
with scale-EMA decay $\delta_s$ and floor $s_{\min}$ ($s^{(0)}$ is initialized to the first observed range). This pins the worst-transferring domain near $0$ and the best near $1$, yielding a bounded, monotone ranking signal. The floored EMA scale guards against a specific failure mode: the cross-domain spread occasionally collapses when all domains' smoothed cosines briefly cluster, and dividing by the raw spread would then produce a one-step spike that, at bandit learning rate $\alpha$, contaminates the Q-values for many subsequent steps.

\subsection{Curriculum Signal and Algorithm}
\label{subsec:tac_rl}
The \alg feedback signal combines the learnability and transferability terms:
\begin{equation}
  S_{m_t}^{(t)}
  \;=\; \beta\,\hat{L}_{m_t}^{(t)} \;+\; (1-\beta)\,T_{m_t}^{(t)},
  \label{eq:combined_score}
\end{equation}
where $\beta\in[0,1]$ interpolates between pure-learnability ($\beta=1$) and pure-transferability ($\beta=0$) selection. The two terms play complementary roles on deliberately different scales: $T_m^{(t)}$ is a bounded relative ranking in $[0,1]$ that orders domains by current cross-domain alignment, while $\hat{L}_{m_t}^{(t)}$ is an unbounded z-score that injects the locally available learning signal; $\beta$ balances the two empirically (\S\ref{subsec:experimental_setup}, Table~\ref{tab:method_hp}).

\vspace{-5pt}
\paragraph{Two-phase value update.}
A standard bandit refreshes only the arm it pulls. But \alg recomputes transferability for \emph{every} domain at each comparison step (Eq.~\ref{eq:minmax_transfer}), so it also has fresh feedback for the arms it did not pull, and uses it. Each step, the sampled arm $m_t$ is updated as usual via Eq.~(\ref{eq:q_update}) with a fresh learnability term and the current $T_{m_t}^{(t)}$. Then, at every comparison step, each \emph{unsampled} arm is updated by the same rule, combining its fresh $T_m^{(t)}$ with its most recent learnability, cached from the last time it was pulled. Arms never yet pulled have no cached learnability and are left untouched. This lets a domain's transferability reallocate sampling mass even while the bandit is not actively revisiting it. We detail the caching and normalizer bookkeeping in Appendix~\ref{app:im_details}.

\vspace{-5pt}
\paragraph{Initialization and warmup.}
With $p_m = |D_m| / \sum_j |D_j|$, Q-values are initialized to centered log-proportions,
\begin{equation}
  Q_m^{(0)} \;=\; \kappa\,\bigl(\log p_m - \tfrac{1}{M}\textstyle\sum_j \log p_j\bigr),
  \label{eq:q_init}
\end{equation}
with scale $\kappa$, so that initial sampling follows $\mu_m^{(0)} \propto |D_m|^{\kappa/\tau}$, proportional to domain size on a softened (log-scaled) footing. In the balanced training setting, where the per-domain cap binds for every domain, this reduces exactly to uniform initialization ($Q_m^{(0)}=0$). At the start of each epoch, the sampler then runs $W$ rounds of round-robin warmup over all $M$ domains (order shuffled per round) before applying Eq.~(\ref{eq:sampling}). Warmup only forces \emph{which} arm is drawn: the bandit, the normalizers, and the visit counts all update live during warmup. This guarantees that every domain contributes multiple gradient-state updates before transferability comparisons drive sampling, and that the learnability normalizer is past its own warmup---with every arm's cached $\hat{L}_m$ refreshed to a properly normalized value---by the time bandit control begins. The UCB bonus in Eq.~(\ref{eq:sampling}) then carries the ongoing exploration burden after warmup, occasionally reviving arms whose Q-values dipped early so that their cached learnability is refreshed.

\vspace{-5pt}
\paragraph{Data-exhaustion policy.}
Each domain maintains an independent index pool per epoch. When the bandit selects a domain whose pool is exhausted, we reshuffle that domain's data and resume sampling from it, rather than re-drawing $m_t$ from the remaining domains. This lets high-$Q$ domains be over-sampled within an epoch while guaranteeing no data is permanently withheld across epochs.

The full procedure, using the bandit template of Eqs.~(\ref{eq:q_update}) and~(\ref{eq:sampling}) with the feedback signal $S_{m_t}^{(t)}$ from Eq.~(\ref{eq:combined_score}), is summarized in Algorithm~\ref{alg:tac}. Additional implementation details (warmup schedule, choice of $N$, $r$, $K_c$) are in Section~\ref{subsec:experimental_setup} and Appendix~\ref{app:im_details}.

\begin{algorithm}[t]
\caption{Transfer-Aware Curriculum for RL (\alg)}
\label{alg:tac}
\begin{algorithmic}[1]
\REQUIRE Domains $\{D_m\}_{m=1}^M$; policy $\pi_\theta$; projection matrix $P$;
         hyperparameters $\alpha, \beta, \gamma, \delta, \delta_s, \kappa, c, \tau, r, N, K_c, W$ (see Table~\ref{tab:method_hp} for values)
\STATE Initialize $\mathbf{h}_m^{(0)}\leftarrow\mathbf{0}$ and $T_m^{(0)}\leftarrow 0$ for all $m$; initialize $Q_m^{(0)}$ via Eq.~(\ref{eq:q_init})
\STATE Run $W$-round round-robin warmup over all $M$ domains (bandit and normalizers update live)
\FOR{step $t = 1, 2, \ldots$}
  \STATE Sample $m_t$ via Eq.~(\ref{eq:sampling}); draw $\mathcal{B}_t\subset D_{m_t}$
  \STATE Compute $K$ rollouts and $\hat{L}_{m_t}^{(t)}$ (Eqs.~\ref{eq:learnability}--\ref{eq:learnability_norm})
  \STATE Compute gradient $g_t$, form unit sketch $\mathbf{v}_t$ (Eq.~\ref{eq:proj_grad}), update $\mathbf{h}_{m_t}^{(t)}$ (Eq.~\ref{eq:grad_ema})
  \STATE \textbf{if} $t \bmod K_c = 0$ \textbf{then} update $\{\rho_m^{(t)}\}_{m \in A_t}$, $\{\tilde{\rho}_m^{(t)}\}_{m \in A_t}$ (Eqs.~\ref{eq:raw_transfer}--\ref{eq:transfer_ema}) and $\{T_m^{(t)}\}_{m \in A_t}$ (Eq.~\ref{eq:minmax_transfer})
  \STATE \textbf{Sampled arm:} compute $S_{m_t}^{(t)}$ (Eq.~\ref{eq:combined_score}); update $Q_{m_t}^{(t)}$ (Eq.~\ref{eq:q_update}); cache $\hat{L}_{m_t}^{(t)}$
  \STATE \textbf{if} $t \bmod K_c = 0$ \textbf{then} update $Q_m^{(t)}$ (Eq.~\ref{eq:q_update}) for each unsampled $m$ with cached $\hat{L}_m$, using $S_m^{(t)} = \beta\hat{L}_m + (1-\beta)\,T_m^{(t)}$
  \STATE Apply one GRPO step on $\mathcal{B}_t$
\ENDFOR
\end{algorithmic}
\end{algorithm}

\vspace{-5pt}
\paragraph{Interpretation.}
\label{subsec:theory}
The gradient-cosine signal admits a local first-order justification under the GRPO surrogate loss minimized at each step, in the spirit of gradient-based influence analyses~\citep{koh2017understanding, pruthi2020estimating, park2023trak, xia2024less}.
A Taylor expansion of $\mathcal{L}_{\mathrm{GRPO}}$ on domain $D_j$ around a step $-\eta g_t$ taken on $D_{m_t}$ gives
\begin{equation}
  \mathcal{L}(\theta - \eta g_t;\,D_j)
  \;\approx\; \mathcal{L}(\theta;\,D_j) \;-\; \eta\,\langle g_t,\,g_j\rangle,
  \label{eq:taylor}
\end{equation}
so $\langle g_t, g_j\rangle$ predicts, to first order, the loss change on $D_j$ from a step on $D_{m_t}$~\citep{pruthi2020estimating}.
The cosine in Eq.~(\ref{eq:raw_transfer}) is the direction-only version of this predictor, mirroring the gradient-alignment view used in multi-task optimization~\citep{yu2020gradient, liu2021conflict}: it preserves the sign of the predicted improvement while normalizing away gradient magnitudes that vary substantially across domains and training stages---a view reinforced by the unit normalization of $\mathbf{v}_t$ in Eq.~(\ref{eq:proj_grad}).
We compute it in a low-dimensional sketch where random projections preserve inner products in expectation~\citep{johnson1984extensions, park2023trak}, and substitute the EMAs $\mathbf{h}_m^{(t)}, \mathbf{h}_j^{(t)}$ for instantaneous gradients to reduce step-level variance.
Averaging over $j\neq m_t$ gives the aggregate cross-domain improvement summarized by $\rho_{m_t}^{(t)}$.
Two properties are worth noting: the signal is \emph{adaptive}, since the gradient states $\{\mathbf{h}_m\}$ evolve with $\pi_\theta$ so $\{T_m^{(t)}\}$ tracks the \emph{current} gradient geometry rather than a static notion of similarity; and \emph{relative}, since the normalization in Eq.~(\ref{eq:minmax_transfer}) rewards a domain for transferring well compared with the other active domains, not for having large gradients.
Together with the learnability term, Eq.~(\ref{eq:combined_score}) thus implements a compact answer to the question \emph{``does a step on $D_m$ both carry local learning signal and point in a direction aligned with the rest of the training set?''}
\vspace{-2pt}
\section{Experiments}
\label{sec:experiments}

\subsection{Experimental Setup}
\label{subsec:experimental_setup}
\vspace{-5pt}
\paragraph{Datasets.}
We train and evaluate on the \textsc{GURU} multi-domain reasoning suite~\citep{chengrevisiting}, spanning six reasoning domains: \textit{math}, \textit{codegen}, \textit{logic}, \textit{simulation}, \textit{table}, and \textit{stem}. We follow the original \textsc{GURU} sources except \textit{stem}, where we replace WebInstruct~\citep{ma2025generalreasoner} with OpenScienceReasoning-2\footnote{\url{https://huggingface.co/datasets/nvidia/OpenScienceReasoning-2}}, a science-reasoning dataset with higher-quality traces. To isolate curriculum design from raw data imbalance, we cap each domain at 1{,}000 queries; without this cap, math and stem dominate training by two orders of magnitude. We additionally construct an \emph{imbalanced} data budget (1{,}500 for math and stem, 500 for simulation and table, 1{,}000 elsewhere) to stress-test the curriculum under realistic source-size skew (\S\ref{subsec:results}). For evaluation, we use held-out benchmarks spanning all six domains: MATH-500~\citep{hendrycks2021measuring}, AIME~\citep{AoPS_AIME} (math); HumanEval~\citep{chen2021evaluating}, MBPP~\citep{austin2021program} (codegen); zebra puzzles~\citep{lin2025zebralogic}, ARC-AGI~\citep{chollet2024arc, chollet2025arc2} (logic); CodeI/O~\citep{li2025codeio}, CruxEval~\citep{gu2024cruxeval} (simulation); HiTab~\citep{cheng2022hitab}, MultiHierTT~\citep{zhao2022multihiertt}, FinQA~\citep{chen2021finqa} (table); GPQA-Diamond~\citep{rein2024gpqa}, SuperGPQA~\citep{du2025supergpqa} (stem). Training data details are in Appendix~\ref{app:datasets}; evaluation pipeline in Appendix~\ref{app:eval_details}; full per-domain prompts in Appendix~\ref{app:prompts}.
\vspace{-5pt}
\paragraph{Models.}
We evaluate on \texttt{Qwen3-1.7B-Base}~\citep{yang2025qwen3} and \texttt{Llama3.2-3B-Instruct}~\citep{grattafiori2024llama}. We use the instruction-tuned Llama variant because its base model produces no correct rollouts on several training domains, collapsing GRPO's advantage signal to zero. Scaling results on \texttt{Qwen3-\{0.6, 4\}B-Base}, confirming \alg's gains hold across model sizes, are in Appendix~\ref{app:model_scaling}.
\vspace{-5pt}
\paragraph{Baselines.}
We compare \alg against four baselines under identical data and compute budgets: \textit{(i)} \textbf{Base}, the model without RL training; \textit{(ii)} \textbf{Random}, which draws each single-domain batch from a domain sampled with fixed probability proportional to its pool size (uniform across domains when pools are balanced); this is the standard pooled multi-domain baseline, with no online adaptation; \textit{(iii)} \textbf{Math-to-Others}~(M2O)~\citep{pang2025reasoning}, a manually designed two-stage curriculum that first trains on math and then on the remaining domains; and \textit{(iv)} \textbf{SEC}~\citep{chen2025self}, an advantage-based learnability curriculum corresponding to \alg with $\beta=1$. Math-to-Others tests hand-designed vs.\ adaptive scheduling; SEC isolates the contribution of transferability.
\vspace{-5pt}
\paragraph{Implementation Details.}
We train all methods with GRPO using $K=4$ rollouts per query, a per-step batch size of 64 queries from a single domain, and 2 epochs total. The bandit uses $\alpha=0.3$, $c=0.2$, softmax temperature $\tau=0.85$, and five rounds of round-robin warmup per epoch ($W=5$). For the transferability signal, we sketch gradients with a Rademacher JL projection~\citep{park2023trak} of dimension $r=4096$ on the last $N=4$ transformer layers, taken from the final PPO mini-batch of each step; we update per-domain gradient EMAs with $\gamma=0.8$, recompute pairwise cosines every $K_c=2$ steps, temporally smooth them with $\delta=0.8$, and map them to per-arm scores via the min--max cross-domain normalization of Eq.~(\ref{eq:minmax_transfer}) (scale-EMA decay $\delta_s=0.9$, floor $s_{\min}=0.01$). The learnability and transferability terms are mixed with $\beta=0.2$. Full tables, compute details, and complexity analysis are in Appendix~\ref{app:training_details} and Appendix~\ref{app:complexity}.

\subsection{Results}
\label{subsec:results}

\begin{table*}[t]
\caption{Pass@1 accuracy (\%) across 14 held-out benchmarks. \textit{Base} is the off-the-shelf model; \textit{Random}, 	\textit{M2O}, and \textit{SEC} are curriculum baselines. Each entry is mean$\pm$std over three training seeds, with most benchmarks additionally averaged over 4 evaluation runs.}
\label{tab:main_results}
\centering
\setlength{\tabcolsep}{3pt}
\renewcommand{\arraystretch}{1.15}
\resizebox{\textwidth}{!}{
\begin{tabular}{l l c c c c c c c c c c}
\toprule
 &  & \multicolumn{5}{c}{\textbf{Qwen3-1.7B}} & \multicolumn{5}{c}{\textbf{Llama3.2-3B}} \\
\cmidrule(lr){3-7} \cmidrule(lr){8-12}
\textbf{Domain} & \textbf{Benchmark} & Base & Random & M2O & SEC & \cellcolor{blue!10}\textbf{TAC} & Base & Random & M2O & SEC & \cellcolor{blue!10}\textbf{TAC} \\
\midrule
\multirow{2}{*}{\textbf{Codegen}} & HumanEval & $27.7_{\pm 0.3}$ & $65.2_{\pm 1.9}$ & $65.0_{\pm 1.8}$ & $64.8_{\pm 0.3}$ & \cellcolor{blue!10}$\mathbf{66.7}_{\pm 1.3}$ & $57.0_{\pm 1.2}$ & $57.0_{\pm 3.7}$ & $54.6_{\pm 2.8}$ & $54.6_{\pm 2.8}$ & \cellcolor{blue!10}$\mathbf{58.4}_{\pm 2.0}$ \\[2pt]
 & MBPP & $36.9_{\pm 0.1}$ & $52.3_{\pm 1.3}$ & $52.0_{\pm 0.8}$ & $52.0_{\pm 1.0}$ & \cellcolor{blue!10}$\mathbf{52.7}_{\pm 2.4}$ & $52.7_{\pm 0.3}$ & $50.5_{\pm 0.7}$ & $51.5_{\pm 0.6}$ & $51.1_{\pm 1.3}$ & \cellcolor{blue!10}$\mathbf{52.8}_{\pm 0.6}$ \\
\cmidrule(lr){1-12}
\multirow{2}{*}{\textbf{Logic}} & ARC-AGI & $0.1_{\pm 0.0}$ & $1.0_{\pm 0.4}$ & $0.6_{\pm 0.1}$ & $0.5_{\pm 0.2}$ & \cellcolor{blue!10}$\mathbf{1.4}_{\pm 0.8}$ & $1.0_{\pm 0.0}$ & $0.8_{\pm 0.7}$ & $0.4_{\pm 0.3}$ & $0.1_{\pm 0.1}$ & \cellcolor{blue!10}$\mathbf{1.2}_{\pm 0.1}$ \\[2pt]
 & Zebra & $1.1_{\pm 0.1}$ & $24.9_{\pm 6.5}$ & $28.5_{\pm 1.4}$ & $26.7_{\pm 3.1}$ & \cellcolor{blue!10}$\mathbf{34.7}_{\pm 0.9}$ & $1.4_{\pm 0.3}$ & $30.5_{\pm 3.7}$ & $30.9_{\pm 2.0}$ & $\mathbf{36.5}_{\pm 1.1}$ & \cellcolor{blue!10}$35.8_{\pm 0.6}$ \\
\cmidrule(lr){1-12}
\multirow{2}{*}{\textbf{Math}} & MATH & $52.8_{\pm 0.8}$ & $59.6_{\pm 1.1}$ & $59.8_{\pm 0.8}$ & $59.1_{\pm 0.5}$ & \cellcolor{blue!10}$\mathbf{60.1}_{\pm 0.6}$ & $34.4_{\pm 0.6}$ & $41.9_{\pm 3.3}$ & $46.1_{\pm 2.0}$ & $43.1_{\pm 2.8}$ & \cellcolor{blue!10}$\mathbf{46.9}_{\pm 0.8}$ \\[2pt]
 & AIME & $4.3_{\pm 0.1}$ & $5.3_{\pm 1.0}$ & $5.2_{\pm 0.4}$ & $\mathbf{5.4}_{\pm 0.6}$ & \cellcolor{blue!10}$5.1_{\pm 0.6}$ & $2.2_{\pm 0.1}$ & $3.4_{\pm 0.8}$ & $4.9_{\pm 1.0}$ & $3.6_{\pm 0.3}$ & \cellcolor{blue!10}$\mathbf{5.2}_{\pm 0.9}$ \\
\cmidrule(lr){1-12}
\multirow{3}{*}{\textbf{Simulation}} & CodeI/O & $3.2_{\pm 0.3}$ & $\mathbf{5.2}_{\pm 0.6}$ & $4.2_{\pm 0.3}$ & $4.4_{\pm 0.6}$ & \cellcolor{blue!10}$4.8_{\pm 0.6}$ & $2.3_{\pm 0.3}$ & $\mathbf{6.7}_{\pm 3.3}$ & $2.0_{\pm 0.7}$ & $2.7_{\pm 1.1}$ & \cellcolor{blue!10}$3.3_{\pm 0.1}$ \\[2pt]
 & CruxEval-I & $15.1_{\pm 0.2}$ & $37.8_{\pm 2.8}$ & $42.3_{\pm 1.8}$ & $\mathbf{42.9}_{\pm 0.4}$ & \cellcolor{blue!10}$42.2_{\pm 2.1}$ & $31.4_{\pm 0.3}$ & $39.7_{\pm 3.4}$ & $36.1_{\pm 0.7}$ & $35.8_{\pm 3.9}$ & \cellcolor{blue!10}$\mathbf{40.6}_{\pm 2.5}$ \\[2pt]
 & CruxEval-O & $15.0_{\pm 0.2}$ & $34.8_{\pm 5.3}$ & $\mathbf{41.5}_{\pm 0.9}$ & $35.8_{\pm 4.0}$ & \cellcolor{blue!10}$39.5_{\pm 3.6}$ & $27.6_{\pm 0.5}$ & $36.2_{\pm 1.7}$ & $\mathbf{37.4}_{\pm 4.6}$ & $34.0_{\pm 4.6}$ & \cellcolor{blue!10}$35.5_{\pm 4.6}$ \\
\cmidrule(lr){1-12}
\multirow{2}{*}{\textbf{STEM}} & GPQA & $24.2_{\pm 0.3}$ & $30.0_{\pm 2.8}$ & $25.8_{\pm 1.1}$ & $30.7_{\pm 1.2}$ & \cellcolor{blue!10}$\mathbf{31.9}_{\pm 0.5}$ & $23.6_{\pm 0.5}$ & $28.0_{\pm 0.8}$ & $27.7_{\pm 0.6}$ & $26.2_{\pm 0.5}$ & \cellcolor{blue!10}$\mathbf{28.7}_{\pm 3.2}$ \\[2pt]
 & SuperGPQA & $18.5_{\pm 0.2}$ & $21.9_{\pm 0.9}$ & $21.7_{\pm 0.5}$ & $21.9_{\pm 0.9}$ & \cellcolor{blue!10}$\mathbf{22.1}_{\pm 0.5}$ & $21.0_{\pm 0.4}$ & $18.8_{\pm 1.3}$ & $\mathbf{21.5}_{\pm 0.8}$ & $18.0_{\pm 1.7}$ & \cellcolor{blue!10}$21.3_{\pm 0.5}$ \\
\cmidrule(lr){1-12}
\multirow{3}{*}{\textbf{Table}} & MultiHierTT & $14.7_{\pm 0.1}$ & $25.5_{\pm 1.2}$ & $24.8_{\pm 0.4}$ & $26.5_{\pm 0.8}$ & \cellcolor{blue!10}$\mathbf{28.7}_{\pm 0.9}$ & $16.6_{\pm 0.4}$ & $21.3_{\pm 2.1}$ & $23.0_{\pm 0.3}$ & $17.8_{\pm 4.8}$ & \cellcolor{blue!10}$\mathbf{24.8}_{\pm 2.0}$ \\[2pt]
 & FinQA & $9.7_{\pm 0.3}$ & $24.3_{\pm 2.5}$ & $24.7_{\pm 0.8}$ & $23.2_{\pm 1.3}$ & \cellcolor{blue!10}$\mathbf{25.6}_{\pm 1.0}$ & $15.6_{\pm 0.3}$ & $14.2_{\pm 1.6}$ & $19.2_{\pm 3.3}$ & $15.4_{\pm 8.2}$ & \cellcolor{blue!10}$\mathbf{21.7}_{\pm 1.0}$ \\[2pt]
 & HiTab & $13.7_{\pm 0.3}$ & $54.5_{\pm 0.6}$ & $51.5_{\pm 2.1}$ & $53.4_{\pm 2.0}$ & \cellcolor{blue!10}$\mathbf{56.5}_{\pm 1.6}$ & $16.6_{\pm 0.3}$ & $60.5_{\pm 1.6}$ & $60.0_{\pm 1.3}$ & $56.5_{\pm 1.9}$ & \cellcolor{blue!10}$\mathbf{61.3}_{\pm 0.6}$ \\
\midrule
\multicolumn{2}{l}{\textbf{\textit{All (macro avg.)}}} & $17.8_{\pm 0.1}$ & $31.8_{\pm 0.5}$ & $32.0_{\pm 0.2}$ & $32.1_{\pm 0.6}$ & \cellcolor{blue!10}$\mathbf{33.9}_{\pm 0.2}$ & $22.2_{\pm 0.2}$ & $29.2_{\pm 0.2}$ & $29.7_{\pm 0.3}$ & $28.5_{\pm 1.5}$ & \cellcolor{blue!10}$\mathbf{31.3}_{\pm 0.6}$ \\
\bottomrule
\end{tabular}
}
\end{table*}

\paragraph{Main Results.}
Table~\ref{tab:main_results} compares \alg with the baselines across both backbones. \alg achieves the best macro-averaged accuracy on both models, improving over the strongest baseline by $+1.8$ pp ($+5.6\%$ relative) on \texttt{Qwen3-1.7B} (over SEC) and $+1.6$ pp ($+5.4\%$ relative) on \texttt{Llama3.2-3B} (over M2O), and ranking first on 10/14 benchmarks on both.

The two adaptive baselines behave inconsistently. M2O improves over Random on both backbones ($+0.2$ on Qwen, $+0.5$ on Llama), confirming that prior domain knowledge yields a useful schedule, but its static two-stage structure cannot react to mid-training shifts in cross-domain transfer. SEC is far less stable: it is the strongest baseline on \texttt{Qwen3-1.7B} ($+0.3$ over Random) yet \emph{underperforms} Random on \texttt{Llama3.2-3B} ($-0.7$), and carries the highest macro variance of any method ($\pm 1.5$ on Llama). As the Curriculum Dynamics analysis below shows, a learnability-only bandit anchors onto whichever domain produces the strongest early advantage signal, regardless of whether its updates benefit the rest of the training set, a gamble that pays off on one backbone but not the other. \alg's transferability term avoids this trap.

\begin{figure}[t]
\centering
\begin{subfigure}[t]{0.32\textwidth}
    \centering
    \includegraphics[width=\textwidth]{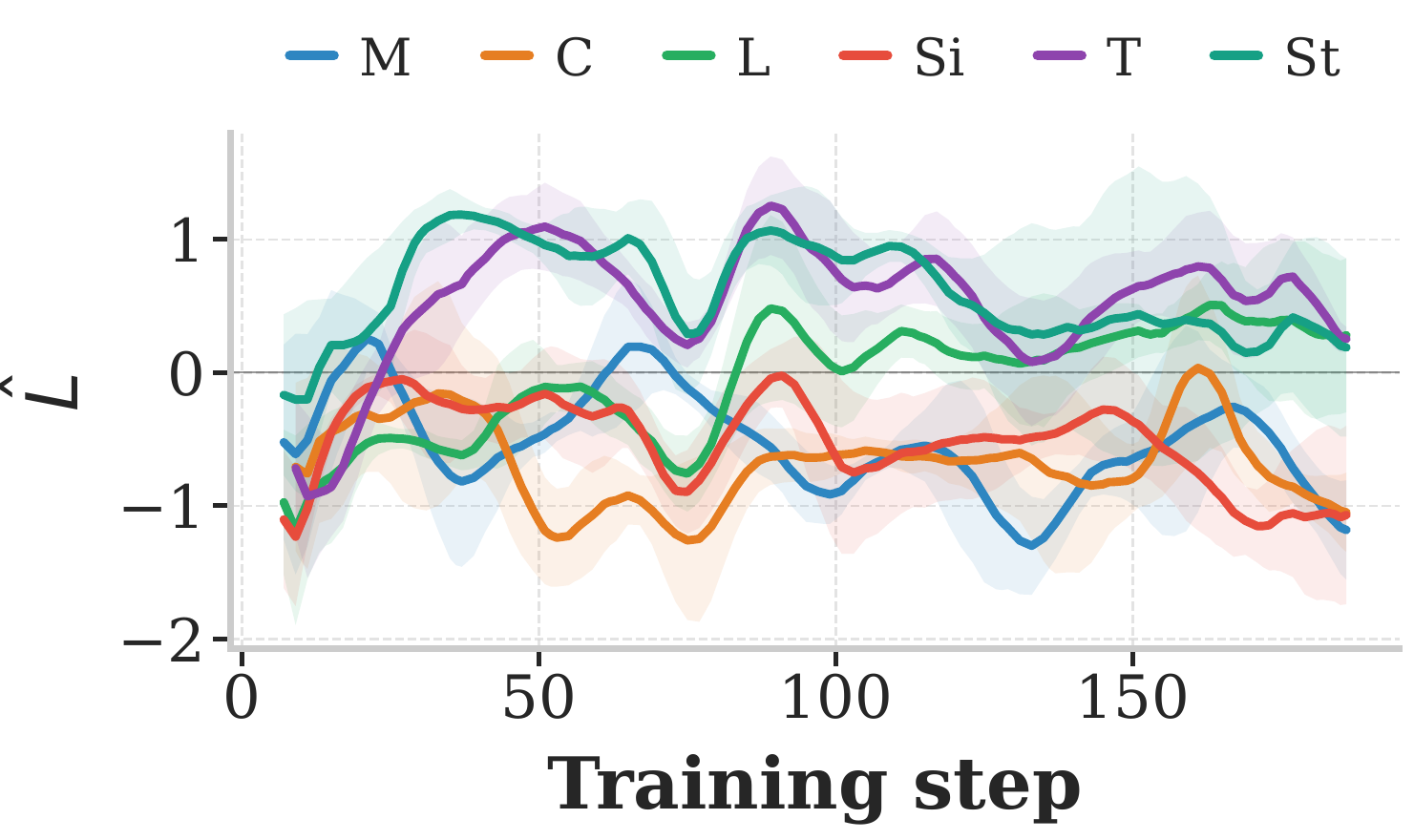}
    \caption{Learnability $\hat{L}_m^{(t)}$.}
    \label{fig:lhat_per_domain}
\end{subfigure}
\hfill
\begin{subfigure}[t]{0.32\textwidth}
    \centering
    \includegraphics[width=\textwidth]{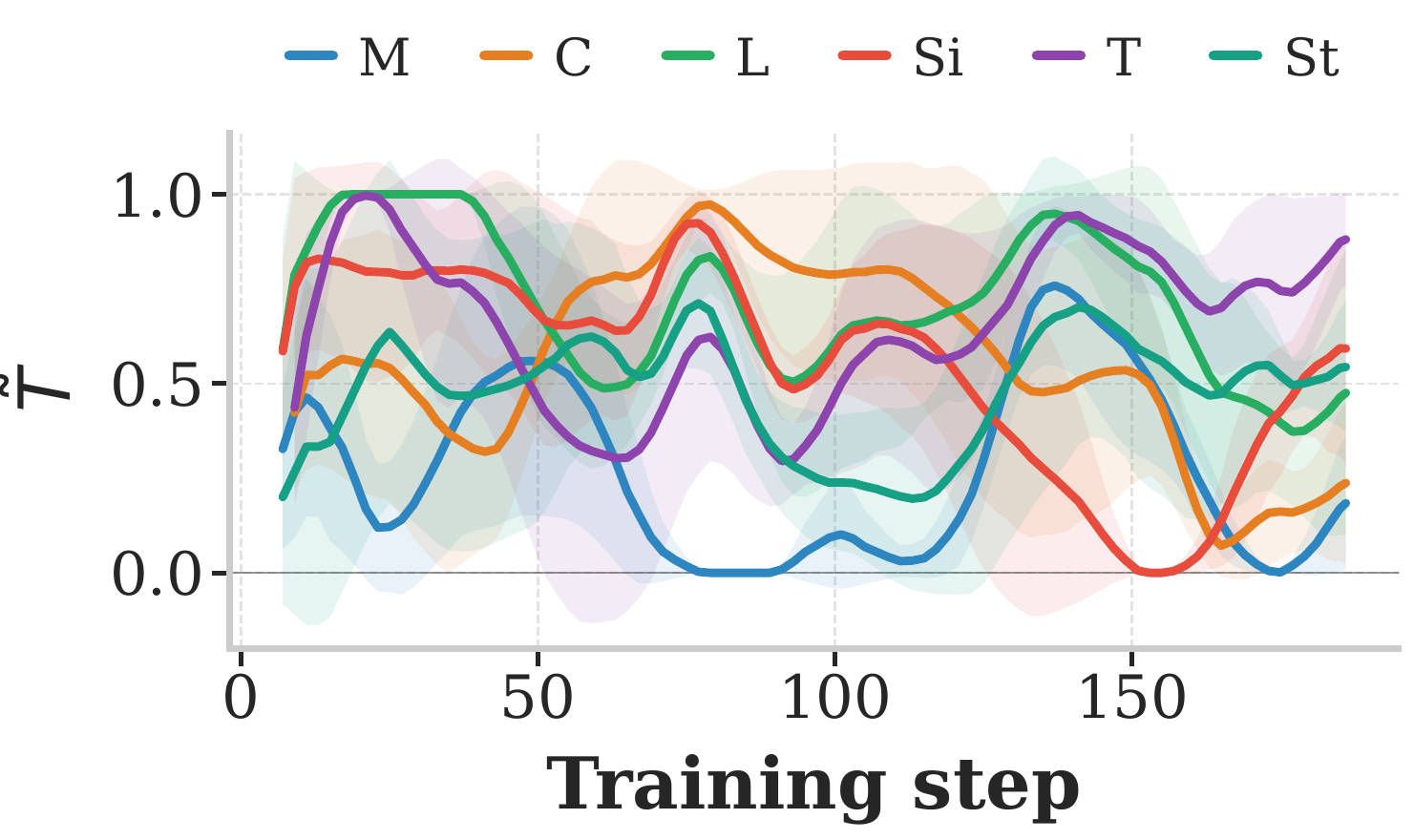}
    \caption{Transferability $T_m^{(t)}$.}
    \label{fig:ttilde_per_domain}
\end{subfigure}
\hfill
\begin{subfigure}[t]{0.32\textwidth}
    \centering
    \includegraphics[width=\textwidth]{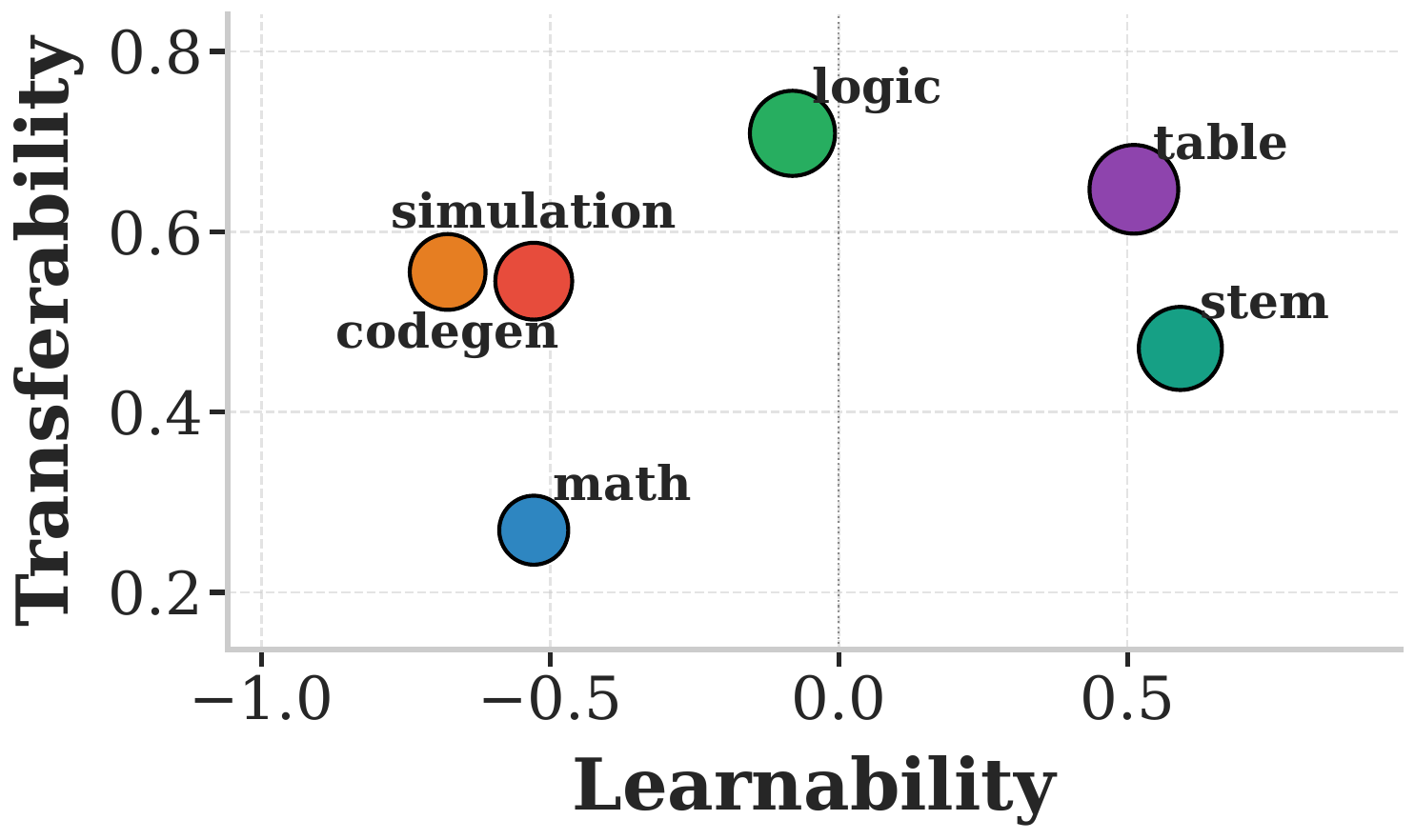}
    \caption{Training-step average of $(\hat{L}_m, T_m)$.}
    \label{fig:lt_alignment}
\end{subfigure}
\caption{\textbf{Learnability and transferability favor different domains.} \textbf{(a)} Per-domain learnability $\hat{L}_m^{(t)}$ across training under \alg, recovered from the bandit's composite score by inverting the $\beta$-mixture $S_m=\beta\,\hat{L}_m+(1-\beta)\,T_m$ for $\hat{L}_m$ (the normalized mean $|A|$ of Eq.~(\ref{eq:learnability})), so that \textbf{(a)} and \textbf{(b)} are measured under the \emph{same} \alg curriculum. \textbf{(b)} Per-domain transferability $T_m^{(t)}$ under \alg, computed via Eqs.~(\ref{eq:raw_transfer})--(\ref{eq:minmax_transfer}). \textbf{(c)} $(\hat{L}_m, T_m)$ per domain, averaged over \emph{all} training steps; marker area is proportional to each domain's mean sampling probability under \alg. Off-axis placement shows the two signals rank domains differently. All curves and markers are averaged over 3 seeds.}
\label{fig:online_signals}
\end{figure}

The gains are broad rather than concentrated. \alg's largest per-benchmark improvements span logic, math, and table reasoning: Zebra rising $+9.8$ over Random and $+8.0$ over SEC on Qwen, MATH rising $+5.0$ over Random on Llama, and FinQA $+7.5$ over Random with MultiHier $+3.5$ on Llama ($+3.2$ on Qwen). They also extend to stem ($+1.9$ on GPQA over Random for Qwen), indicating the transferability signal is not biased toward any particular target. The curriculum analysis flags \emph{math} and \emph{codegen} as the \emph{least} transferable domains: their training gradients align weakly, even negatively, with every other domain (Appendix~\ref{app:pairwise}), an effect we attribute to base models such as \texttt{Qwen3} being heavily pretrained on math and code. Yet \alg, which \emph{down-weights} both relative to uniform sampling, still improves their benchmarks. We read this as capability-level spillover, where investing the schedule in the broadly-transferable domains reinforces general verifiable-task solving and exposes non-math reasoning styles that carry over to math and code at evaluation, even though the local gradient-cosine signal does not directly credit it. The only benchmarks where \alg is not first on \emph{either} backbone, CodeI/O and CruxEval-O, both fall in the simulation domain, where local learnability and transferability are both relatively flat (Figure~\ref{fig:online_signals}) and the curriculum has limited leverage. The gains also transfer across families: although \texttt{Llama3.2-3B} starts substantially weaker than \texttt{Qwen3-1.7B} (e.g., 34.4 vs.\ 52.8 on MATH), \alg improves over Random by an identical $+2.1$ macro points on both backbones, which we attribute to the transferability signal being computed from each model's own training gradients. Examples of the resulting qualitative differences in model rollouts are provided in Appendix~\ref{app:more_examples}.

\begin{figure}[t]
\centering
\begin{subfigure}[t]{0.32\textwidth}
    \centering
    \includegraphics[width=\textwidth]{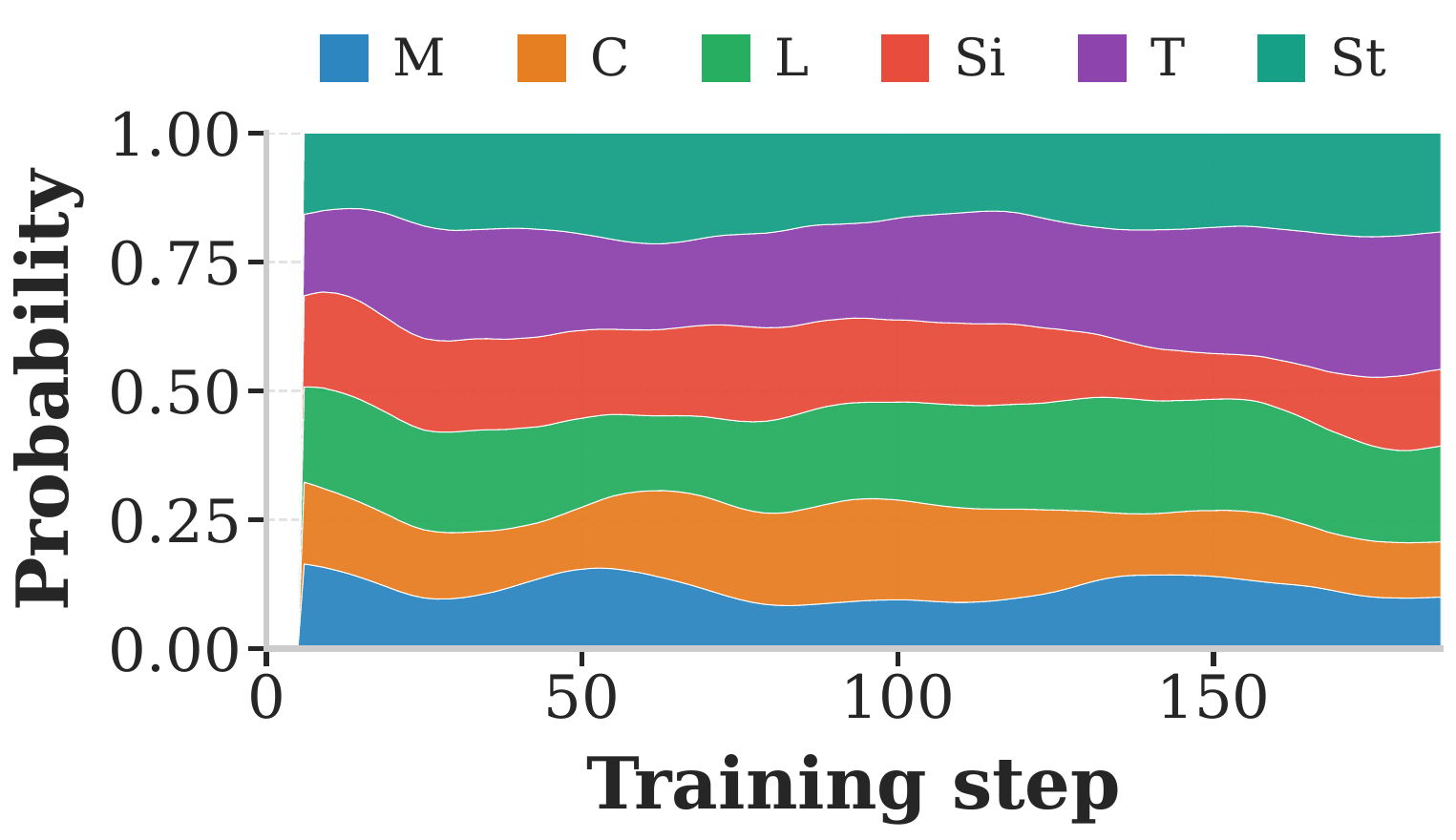}
    \caption{Sampling under \alg.}
    \label{fig:sampling_tac}
\end{subfigure}
\hfill
\begin{subfigure}[t]{0.32\textwidth}
    \centering
    \includegraphics[width=\textwidth]{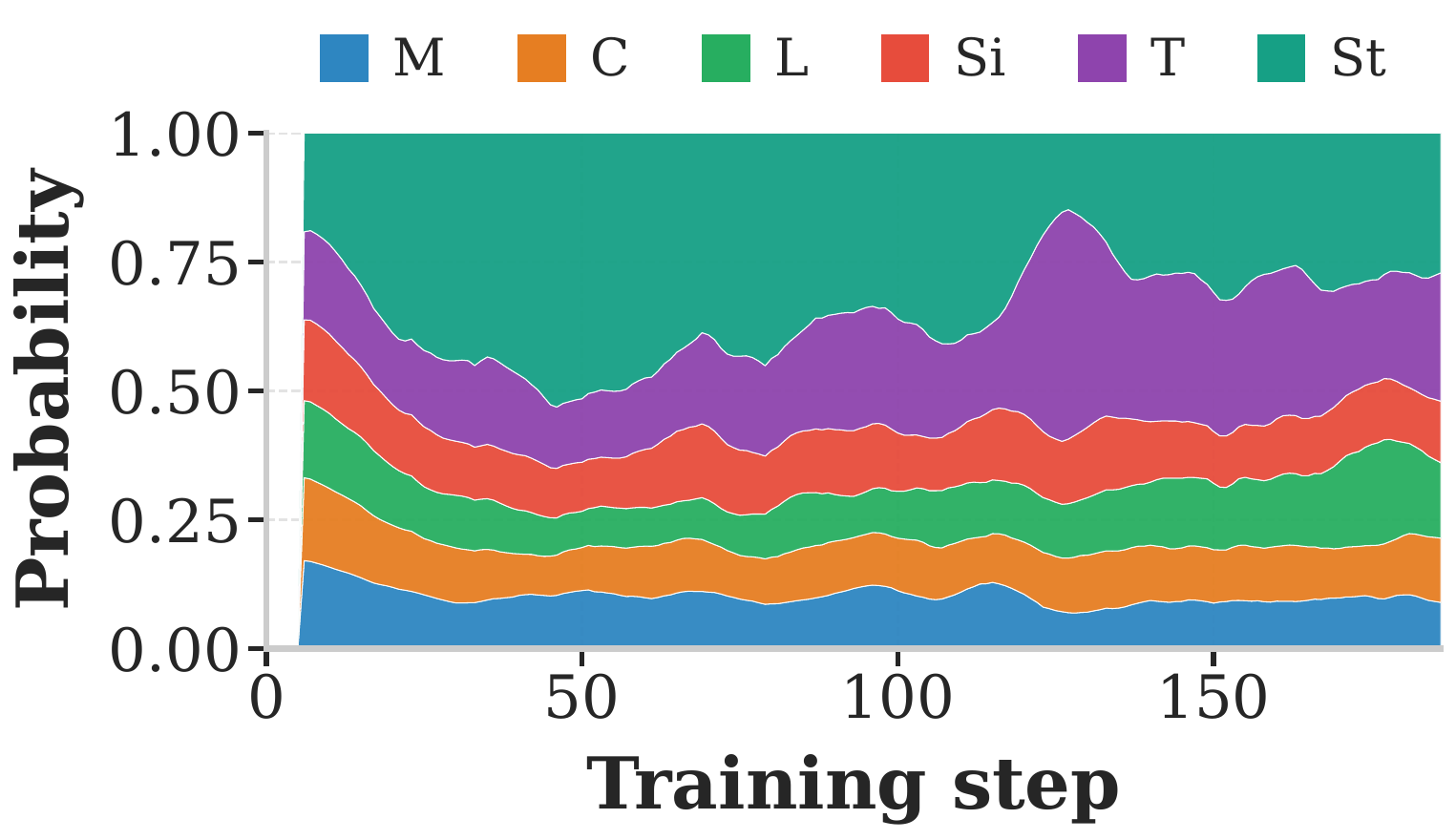}
    \caption{Sampling under SEC ($\beta\!=\!1$).}
    \label{fig:sampling_sec}
\end{subfigure}
\hfill
\begin{subfigure}[t]{0.32\textwidth}
    \centering
    \includegraphics[width=\textwidth]{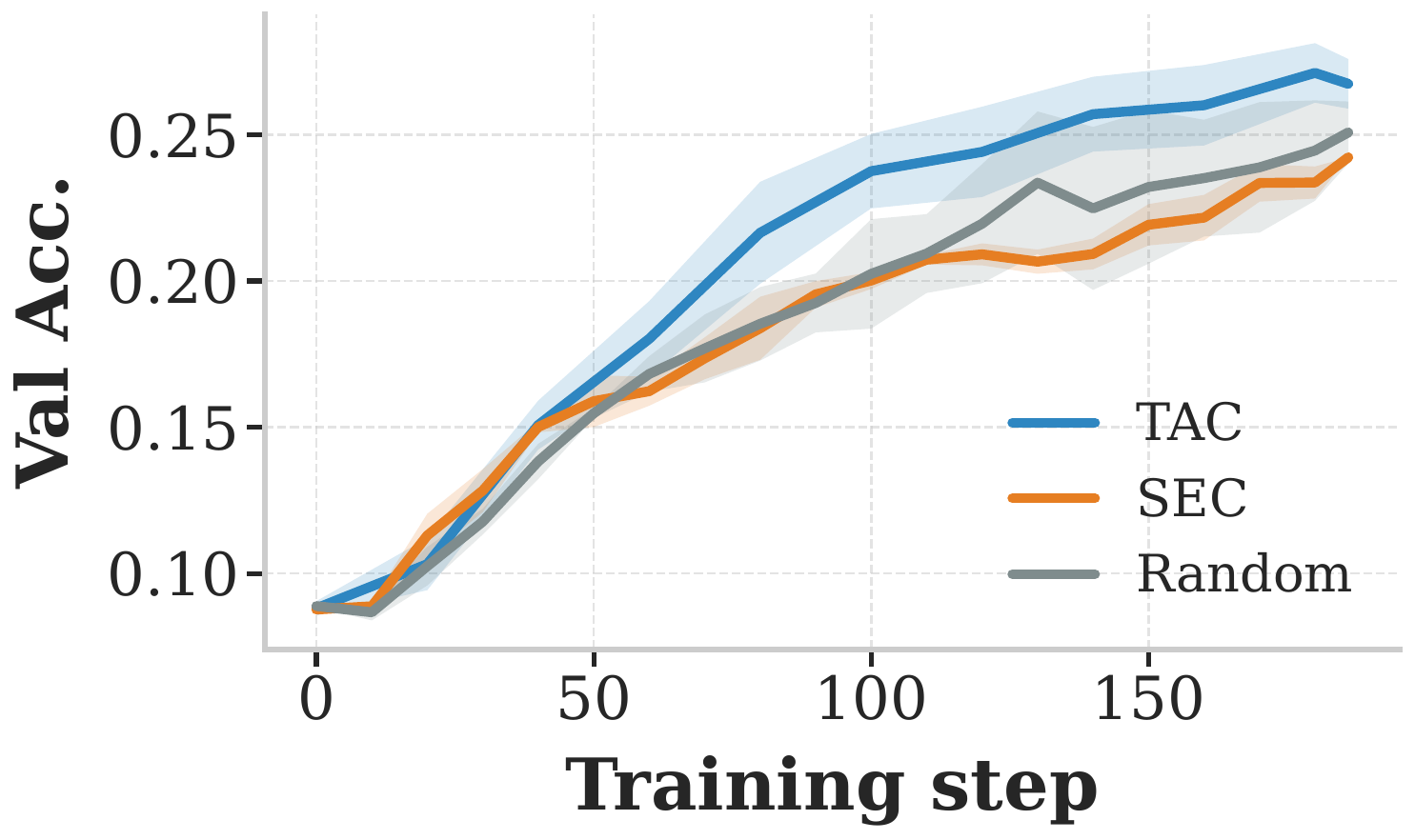}
    \caption{Macro validation accuracy.}
    \label{fig:val_curve}
\end{subfigure}
\caption{\textbf{Curriculum dynamics.} \textbf{(a)--(b)} Per-domain sampling probability $\mu_m^{(t)}$ under \alg and SEC; colors index the six training domains (\textit{math, codegen, logic, simulation, table, stem}). \textbf{(c)} Macro-averaged validation accuracy across all evaluation benchmarks.}
\label{fig:curriculum_dynamics}
\vspace{-10pt}
\end{figure}
\vspace{-5pt}
\paragraph{Curriculum Dynamics: \alg Adapts; SEC Anchors.}
Because the curriculum is online, \emph{when} each method samples each domain matters as much as what. Both \alg and SEC begin near proportional initialization and behave similarly for $\sim\!30$ steps. Around step $30$--$50$, \emph{stem} learnability spikes (Figure~\ref{fig:lhat_per_domain}, peaking near step $\sim\!35$ before it is trained down) and SEC's bandit collapses onto it: by step $\sim\!60$, stem occupies about half of SEC's sampling mass (Figure~\ref{fig:sampling_sec}) and stays the dominant domain thereafter. \alg up-weights stem only transiently (Figure~\ref{fig:sampling_tac}), peaking near step $\sim\!60$, then releases it as \emph{table} overtakes every other domain in transferability (Figure~\ref{fig:ttilde_per_domain}); \alg redirects the freed budget toward \emph{table}, which reaches its largest sampling share by the end of training. The decisive point is that the most \emph{learnable} domain (stem) and the most \emph{transferable} domain (table) are not the same: SEC, ranking domains by learnability alone, pours most of its mass into stem, whereas \alg, rewarding transferability on top of learnability, shifts the budget onto table (strong on \emph{both} signals) and logic. The validation curves (Figure~\ref{fig:val_curve}) trace the same timeline: the three methods are indistinguishable until step $\sim\!40$, after which \alg pulls ahead while SEC tracks Random closely, indicating that without a transferability-sensitive signal the bandit machinery alone does not improve over uniform sampling.

\vspace{-5pt}
\paragraph{Learnability and Transferability Favor Different Domains.}
Because both signals are read from \alg's own run, learnability recovered as the normalized mean $|A|$ ($\hat{L}$) and transferability as $T$, Figures~\ref{fig:lhat_per_domain} and~\ref{fig:ttilde_per_domain} track $\hat{L}$ and $T$ under a single curriculum, and the domains they favor separate \emph{progressively}. \textbf{Early ($t\!\lesssim\!50$):} learnability spikes for stem (and table) while $T$ is uniformly high and not yet discriminative, so SEC and \alg make similar decisions. \textbf{Mid ($t\!\approx\!50$--$100$):} stem's $\hat{L}$ stays high but begins to fall as it is trained down, table's $\hat{L}$ holds, and $T$ differentiates sharply: table climbs to the highest transferability while \emph{codegen} and \emph{math} drift to the bottom, so \alg's sampling peels away from SEC's toward table. \textbf{Late ($t\!\gtrsim\!100$):} stem's $\hat{L}$ keeps falling while \emph{logic}'s climbs, so the two orderings stay misaligned rather than converging. Averaged over the whole run (Figure~\ref{fig:lt_alignment}, marker area $\propto$ sampling share), this leaves stem at high $\hat{L}$ but only moderate $T$, \emph{logic} at low $\hat{L}$ yet high $T$, table high on \emph{both} axes and sampled most, and \emph{math} low on both and sampled least. A learnability-only curriculum locks onto the high-$\hat{L}$ stem and under-weights the equally- or more-transferable table and logic; \alg uses $T$ to correct this.

This online ranking is consistent with the offline transfer matrix in Figure~\ref{fig:transfer_matrix}, where \emph{table} as a single source yields among the largest off-diagonal accuracy gains and \emph{math} the smallest, though the two need not match exactly: the matrix measures end-to-end accuracy transfer from training each domain in isolation, whereas \alg's signal is an online, first-order estimate of gradient alignment that evolves with the policy. The per-pair gradient cosines in Appendix~\ref{app:pairwise} confirm the same structure: \emph{math} and \emph{codegen} align weakly, often negatively, with every other domain and most weakly with \emph{each other}, while \emph{table} behaves as a hub that aligns positively with simulation, stem, and logic. We attribute the math/codegen isolation to a pretraining effect: base models, and \texttt{Qwen3} in particular, are saturated with math and code, so their RL updates point in idiosyncratic directions that are near-orthogonal to the shared reasoning gradients the other domains move along, sampling them more sharpens an already-strong skill rather than benefiting the rest.

\vspace{-5pt}
\paragraph{Ablation: Components of \alg.}
\begin{figure}[t]
\centering
\begin{subfigure}[t]{0.32\textwidth}
    \centering
    \includegraphics[width=\textwidth]{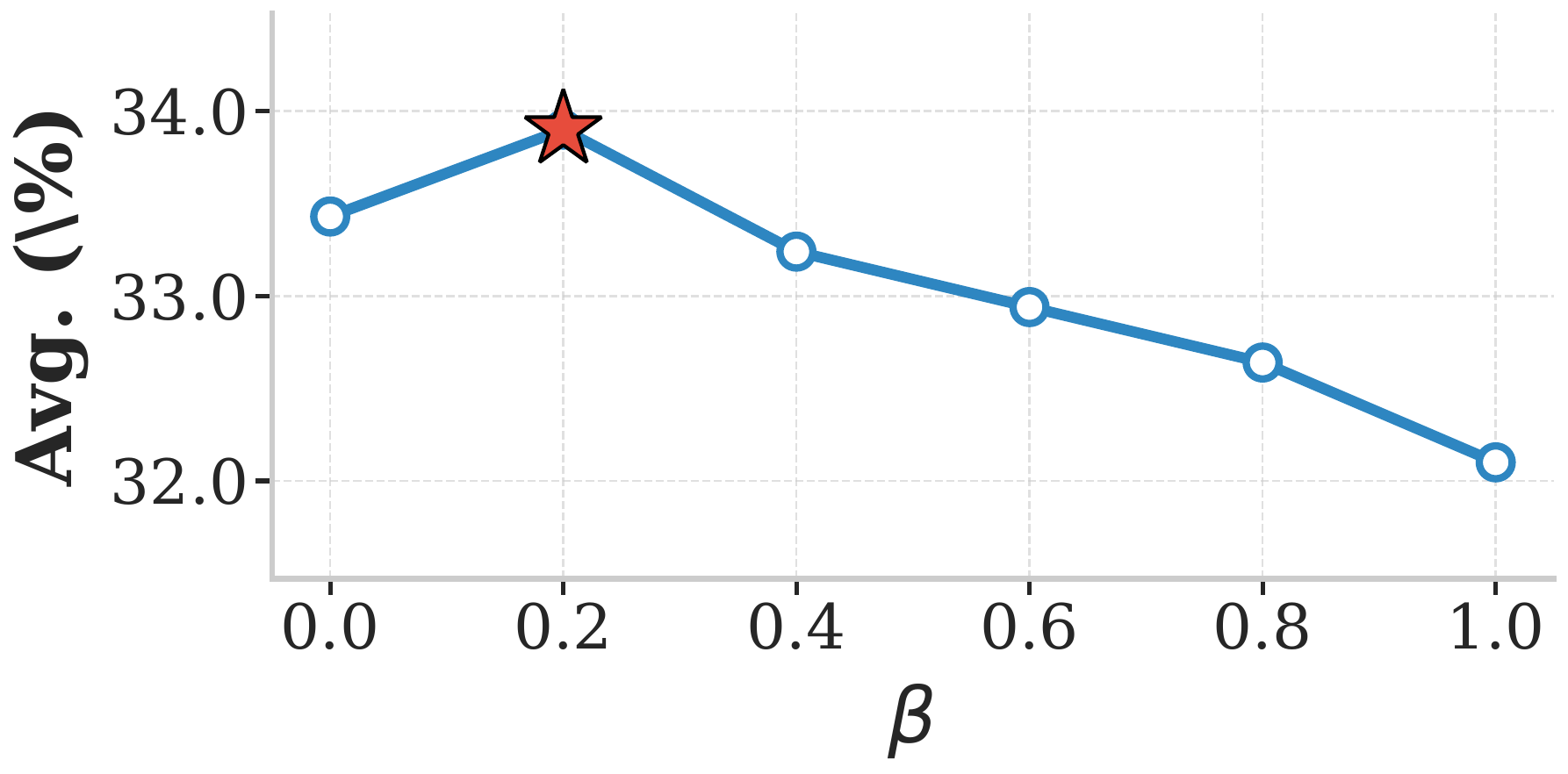}
    \caption{Mixing coefficient $\beta$.}
    \label{fig:ablation_beta}
\end{subfigure}
\hfill
\begin{subfigure}[t]{0.32\textwidth}
    \centering
    \includegraphics[width=\textwidth]{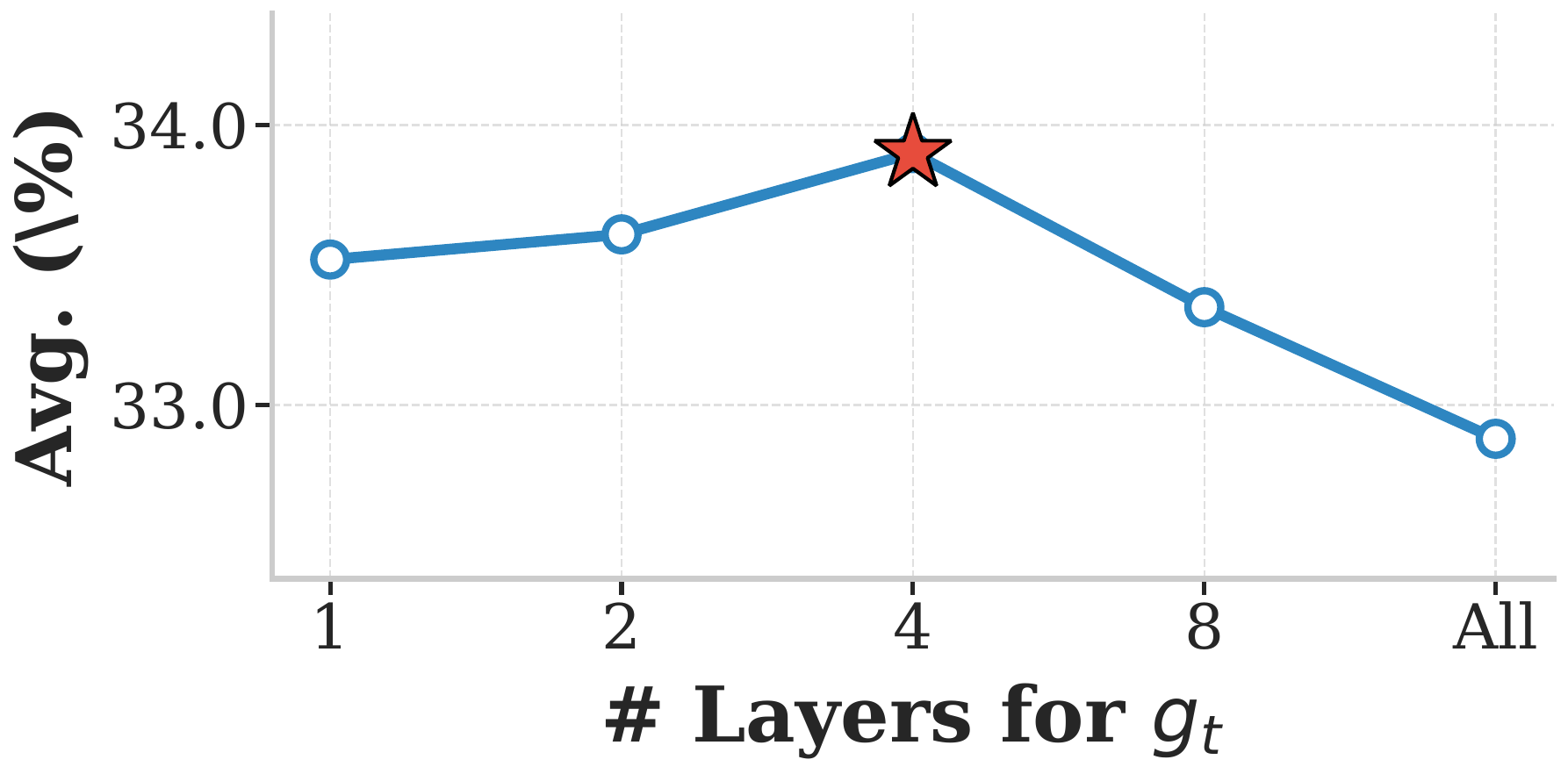}
    \caption{Number of layers $N$.}
    \label{fig:ablation_layers}
\end{subfigure}
\hfill
\begin{subfigure}[t]{0.32\textwidth}
    \centering
    \includegraphics[width=\textwidth]{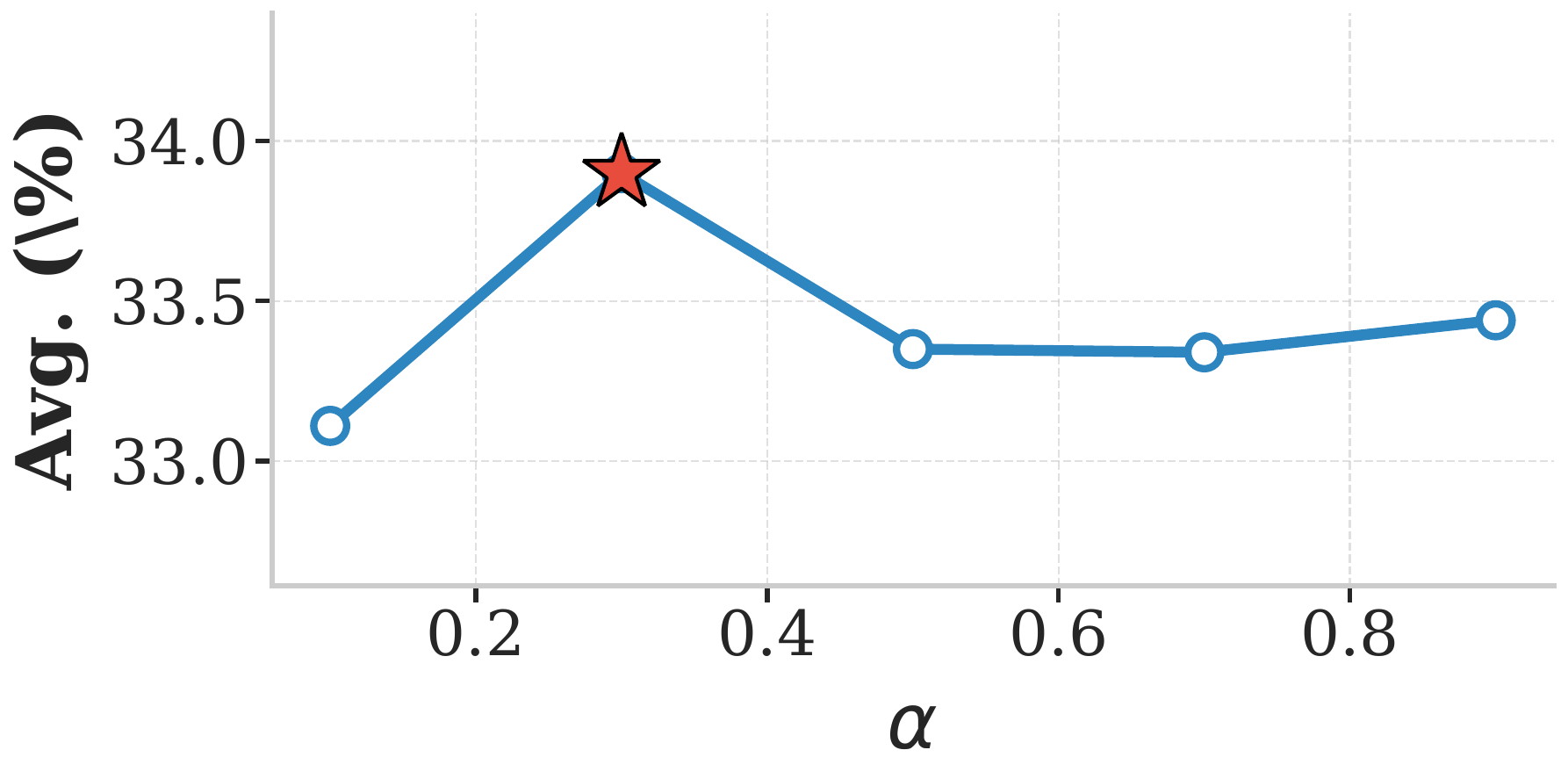}
    \caption{Bandit learning rate $\alpha$.}
    \label{fig:ablation_alpha}
\end{subfigure}
\caption{\textbf{Hyperparameter ablations.} Macro-averaged accuracy (unweighted mean over the six domain scores) while varying \textbf{(a)} the mixing coefficient $\beta$ ($\beta\!=\!0$: pure transferability; $\beta\!=\!1$: pure learnability); \textbf{(b)} the number of trailing transformer layers $N$ used for the projected gradient (Eq.~\ref{eq:proj_grad}); and \textbf{(c)} the bandit EMA learning rate $\alpha$ (Eq.~\ref{eq:q_update}). Shaded regions denote standard deviation over three seeds.}
\label{fig:ablations}
\end{figure}
We ablate \alg's three main hyperparameters: the mixing coefficient $\beta$, the bandit learning rate $\alpha$, and the number of trailing layers $N$ used for the projected gradient. Performance is stable within $\pm 0.2$ points around our default settings (\autoref{fig:ablations}). $\beta$ peaks at $\beta\!\approx\!0.2$ and degrades sharply at $\beta\!=\!1$ (\autoref{fig:ablation_beta}), confirming the contribution of transferability; $\beta\!=\!0$ is competitive but slightly weaker, indicating learnability still adds signal. $\alpha$ peaks at $0.3$ (\autoref{fig:ablation_alpha}): smaller values update too slowly given how fast the gradient geometry changes during RL, larger values amplify step-level noise. $N$ has the smallest effect (\autoref{fig:ablation_layers}): even a few trailing layers already match the full-model gradient, indicating the last layers carry most of the cross-domain alignment signal, consistent with gradient directions in trailing layers being highly correlated. Additional ablations on the projection family, projection dimension $r$, and the data-exhaustion strategy are in Appendix~\ref{app:more_ablations}.

\vspace{-5pt}
\paragraph{Additional Study: Imbalanced Data Budget.}
The main results equalize the data budget across domains to isolate the curriculum's contribution from raw data skew. In practice, training pools are rarely balanced. We re-run training on an imbalanced data budget (1{,}500 for math and stem, 500 for simulation and table, 1{,}000 elsewhere) and evaluate on the same 14 benchmarks; Table~\ref{tab:imbalanced_results} reports results for \texttt{Qwen3-1.7B}.

\alg again leads on macro-averaged accuracy ($32.7$, $+1.9$ over Random, $+1.5$ over SEC), and ranks first on 11 of 14 benchmarks. The gap over SEC is the more informative one: the centered log-proportional Q-initialization gives math and stem roughly twice (about $1.9\times$) the initial mass of simulation and table, and a learnability-only signal provides no corrective force when those large-but-narrow domains saturate. \alg's transferability term supplies that correction, releasing weight from over-represented domains once their updates stop benefiting the rest, visible in the largest gains accruing to under-represented domains (CodeI/O $+2.1$ over Random, FinQA $+4.0$ over SEC). \alg's gains are thus robust to data-budget skew, not an artifact of the balanced setting.
\begin{table*}[t]
\caption{\textbf{Imbalanced data budget (\texttt{Qwen3-1.7B}).} Math and STEM use 1500 training samples; Simulation and Table use 500; remaining domains use 1000. Evaluation benchmarks are unchanged. Mean over three seeds; best per column in bold.}
\label{tab:imbalanced_results}
\centering
\small
\setlength{\tabcolsep}{3pt}
\renewcommand{\arraystretch}{1.10}
\resizebox{\textwidth}{!}{
\begin{tabular}{l c c c c c c c c c c c c c c c}
\toprule
\textbf{Method} & \multicolumn{2}{c}{\textbf{Codegen}} & \multicolumn{2}{c}{\textbf{Logic}} & \multicolumn{2}{c}{\textbf{Math}} & \multicolumn{3}{c}{\textbf{Simulation}} & \multicolumn{2}{c}{\textbf{STEM}} & \multicolumn{3}{c}{\textbf{Table}} & \textbf{All} \\
\cmidrule(lr){2-3} \cmidrule(lr){4-5} \cmidrule(lr){6-7} \cmidrule(lr){8-10} \cmidrule(lr){11-12} \cmidrule(lr){13-15} \cmidrule(lr){16-16}
 & HE & MBPP & ARC-AGI & Zebra & MATH & AIME & CodeI/O & CruxEval-I & CruxEval-O & GPQA & SuperGPQA & MH & FinQA & HiTab & \textit{macro avg.} \\
\midrule
Random & $66.6$ & $\mathbf{53.8}$ & $0.5$ & $12.0$ & $60.8$ & $4.7$ & $4.4$ & $41.1$ & $38.8$ & $30.6$ & $20.3$ & $25.9$ & $23.0$ & $47.6$ & $30.8$ \\
SEC & $64.6$ & $51.7$ & $\mathbf{1.6}$ & $21.2$ & $58.9$ & $4.6$ & $4.2$ & $40.8$ & $36.8$ & $31.9$ & $20.6$ & $25.9$ & $19.7$ & $\mathbf{50.8}$ & $31.2$ \\
\cellcolor{blue!10}\textbf{\alg} & \cellcolor{blue!10}$\mathbf{67.1}$ & \cellcolor{blue!10}$52.7$ & \cellcolor{blue!10}$1.2$ & \cellcolor{blue!10}$\mathbf{23.6}$ & \cellcolor{blue!10}$\mathbf{61.1}$ & \cellcolor{blue!10}$\mathbf{5.9}$ & \cellcolor{blue!10}$\mathbf{6.5}$ & \cellcolor{blue!10}$\mathbf{42.3}$ & \cellcolor{blue!10}$\mathbf{40.0}$ & \cellcolor{blue!10}$\mathbf{33.0}$ & \cellcolor{blue!10}$\mathbf{22.2}$ & \cellcolor{blue!10}$\mathbf{26.5}$ & \cellcolor{blue!10}$\mathbf{23.7}$ & \cellcolor{blue!10}$49.2$ & \cellcolor{blue!10}$\mathbf{32.7}$ \\
\bottomrule
\end{tabular}
}
\end{table*}
\section{Related Work}
\label{sec:related_work}
\vspace{-5pt}
\paragraph{Multi-Domain Reasoning for LLMs.}
Multi-domain training has roots in classical multi-task learning, including shared-representation approaches~\citep{caruana1997multitask, ruder2017overview}, gradient-conflict mitigation~\citep{yu2020gradient, liu2021conflict}, and data-mixture optimization for language model pretraining~\citep{xie2023doremi}. The line we build on extends these ideas to RL training of LLM reasoners across multiple reasoning domains. General Reasoner~\citep{ma2025generalreasoner} and GURU~\citep{chengrevisiting} introduce broad multi-domain reasoning benchmarks, including non-verifiable domains that rely on LLM-as-a-judge rewards~\citep{zheng2023judging}, and show that math-only RL is insufficient for general-purpose reasoning. \citet{huan2025does} and \citet{li2025can} further examine when and how reasoning skills transfer across domains. Building on these benchmarks, methodological work has begun to address multi-domain training: \citet{liang2026boosting} encourage gradient alignment across domains during RL, while \citet{ramesh2026multi} reweight per-task losses for balanced multi-task GRPO. Both modify the optimizer or loss; we instead target the sampling distribution, selecting domains by their joint learnability and cross-domain transferability.
\vspace{-5pt}
\paragraph{Curriculum Learning for RL.}
Curriculum learning has long been studied in RL~\citep{matiisen2019teacher, wang2020learning, wang2021survey, tzannetos2023proximal}, typically by estimating task difficulty or learning progress and then adaptively selecting tasks, goals, or environments to improve sample efficiency. Recent work extends this to RL training of LLM reasoners. Omni-Thinker~\citep{li2025omni} updates its sampling distribution using an accuracy-based backward-transfer signal computed on held-out evaluation probes; \citet{pang2025reasoning} propose a math-to-others curriculum that leverages math reasoning as a foundation for other tasks; and easy-to-hard curricula~\citep{parashar2026curriculum} have likewise been adopted. DUMP~\citep{wang2025dump} and SEC~\citep{chen2025self} use advantage-based learnability to construct curricula automatically, with a similar idea applied to variance-based rewards~\citep{jiang2025vcrl}. \alg shares the bandit-curriculum framework and advantage-based learnability term of these methods; its contribution is a complementary cross-domain transferability signal, estimated directly from training gradients without additional rollouts or held-out probes, that selects domains whose updates are directionally aligned with the rest of the training set.
\section{Conclusion}
\label{sec:conclusion}
We introduce \fullalg~(\alg), a bandit-style curriculum for multi-domain RL that combines local learnability with a gradient-geometry estimate of cross-domain transferability, computed entirely from training gradients and evolving with the policy as training proceeds. Across a six-domain reasoning suite, \alg consistently outperforms proportional sampling, a hand-designed math-to-others curriculum, and a learnability-only bandit on both \texttt{Qwen3-1.7B} and \texttt{Llama3.2-3B-Instruct}, in both balanced and imbalanced training mixtures. We hope these results motivate further work on cross-domain transferability as a first-class signal for curriculum design in multi-domain reasoning. 


\vspace{-5pt}
\paragraph{Limitations.}
\alg targets curriculum design, the sampling distribution over training domains, rather than the optimizer, the reward model, or the policy architecture. This scoping is deliberate, but it also means \alg's gains are complementary to, not substitutes for, advances along those other axes, and the strongest multi-domain reasoning systems will likely combine adaptive curricula with progress on optimization and reward design. Our experiments are likewise restricted to the RL-with-verifiable-rewards (RLVR) regime, where advantage-based learnability and gradient-based transferability admit clean definitions; extending these signals to domains with non-verifiable or model-judged rewards~\citep{zheng2023judging} is left to future work.

\vspace{-5pt}
\paragraph{Broader Impact.}
RL post-training of LLMs has so far concentrated on a narrow set of domains, most prominently mathematics and code generation, where verifiable rewards are easy to construct. \alg makes it cheaper and more reliable to train across a broader portfolio of reasoning domains simultaneously, which we view as a step toward reasoning capability that generalizes beyond the math-and-code corner of the design space, into domains such as scientific reasoning, table understanding, and structured logic. By improving sample efficiency rather than introducing new data sources or deployment capabilities, \alg also lowers the compute barrier to training competent reasoning models, broadening access for academic and resource-constrained developers. We do not anticipate domain-specific harms beyond those already associated with RL fine-tuning of large language models.

\section*{Acknowledgement}

We especially thank Roger Grosse, Minsu Kim, and Kimin Lee for providing important suggestions on this project. We also thank Ryan Faulkner for discussions on the RL setup. Finally, we sincerely thank Jimin Lee for her help with the paper and figures.

This work was supported in part by the German Federal Ministry of Education and Research (BMBF): Tübingen AI Center, FKZ: 01IS18039B; by the Machine Learning Cluster of Excellence, EXC number 2064/1 – Project number 390727645; 
by the NSERC Discovery Grant RGPIN-2025-06491; 
and by the University of Toronto’s Acceleration Consortium, which receives funding from the Canada First Research Excellence Fund (CFREF).
Resources used in preparing this research project were provided, in part, by the Digital Research Alliance of Canada; the Province of Ontario; the Government of Canada through CIFAR; and companies sponsoring the Vector Institute.

\clearpage
\bibliography{reference}
\bibliographystyle{plainnat}


\clearpage
\appendix
\definecolor{institution-bg}{HTML}{E6F2F5}
\definecolor{institution-frame}{HTML}{2C88A0}
\definecolor{contribution-bg}{HTML}{E6F5E9}
\definecolor{contribution-frame}{HTML}{2E8B57}
\definecolor{punishment-bg}{HTML}{F0E6F5}
\definecolor{punishment-frame}{HTML}{8A4F9E}
\definecolor{anonymous-bg}{HTML}{F2F2F2}
\definecolor{anonymous-frame}{HTML}{A9A9A9}
\definecolor{classification-bg}{HTML}{FFF6E6}
\definecolor{classification-frame}{HTML}{D4A76A}
\definecolor{evaluator-bg}{rgb}{0.97,0.95,1}     
\definecolor{evaluator-frame}{rgb}{0.5,0.4,0.7}  

\definecolor{system-bg}{rgb}{0.95,0.95,1} 
\definecolor{system-frame}{rgb}{0.4,0.4,0.8}  
\definecolor{user-bg}{rgb}{0.95,1,0.95}   
\definecolor{user-frame}{rgb}{0.4,0.7,0.4}  
\definecolor{assistant-bg}{rgb}{1,0.95,0.95} 
\definecolor{assistant-frame}{rgb}{0.8,0.4,0.4}  
\definecolor{institution-bg}{rgb}{0.98,0.98,0.98} 
\definecolor{institution-frame}{rgb}{0.6,0.6,0.6} 

{
    \newpage
    \centering
    \Large
    \vspace{1.0em}
    \textit{\textbf{\mytitle}} \\
    \vspace{0.5em}Supplementary Material \\
    \vspace{1.0em}
}

\section{Experiment Prompts}
\label{app:prompts}

For all experiments, we use the default prompt templates from the VERL\footnote{\url{https://github.com/volcengine/verl}} ~\citep{sheng2025hybridflow} repository.

\subsection{Math}

\begin{tcolorbox}[enhanced, breakable, title={User Prompt (Math)}, colback=user-bg, colframe=user-frame, colbacktitle=user-frame!30!white, coltitle=black, fonttitle=\bfseries, boxrule=0.5mm, arc=2mm]
\begin{verbatim}
{{question}} Please output the final answer within \boxed{}.
\end{verbatim}
\end{tcolorbox}

\subsection{Code Generation}

\paragraph{LeetCode2K.} The user prompt is the cleaned dataset query (after removing \verb|### Answer:| and normalizing \verb|### Format:|):

\begin{tcolorbox}[enhanced, breakable, title={User Prompt (LeetCode2K)}, colback=user-bg, colframe=user-frame, colbacktitle=user-frame!30!white, coltitle=black, fonttitle=\bfseries, boxrule=0.5mm, arc=2mm]
\begin{verbatim}
{{query}}
\end{verbatim}
\end{tcolorbox}

\paragraph{LiveCodeBench (Function Call).}

\begin{tcolorbox}[enhanced, breakable, title={User Prompt (LiveCodeBench, Function Call)}, colback=user-bg, colframe=user-frame, colbacktitle=user-frame!30!white, coltitle=black, fonttitle=\bfseries, boxrule=0.5mm, arc=2mm]
\begin{verbatim}
You are an expert Python programmer. You will be given a question
(problem specification) and will generate a correct Python program
that matches the specification and passes all tests.

Below is the question:

{{problem_desc}}

You will use the following starter code to write the solution to the
problem and enclose your code within ```python delimiters.
```python
{{starter_code}}
```
\end{verbatim}
\end{tcolorbox}

\paragraph{LiveCodeBench (STDIN).}

\begin{tcolorbox}[enhanced, breakable, title={User Prompt (LiveCodeBench, STDIN)}, colback=user-bg, colframe=user-frame, colbacktitle=user-frame!30!white, coltitle=black, fonttitle=\bfseries, boxrule=0.5mm, arc=2mm]
\begin{verbatim}
You are an expert Python programmer. You will be given a question
(problem specification) and will generate a correct Python program
that matches the specification and passes all tests.

Below is the question:

{{problem_desc}}

Read the inputs from stdin, solve the problem, and write the answer to
stdout (do not directly test on the sample inputs). Enclose your code
within delimiters. Ensure that when the python program runs, it reads
the inputs, runs the algorithm, and writes output to STDOUT.
\end{verbatim}
\end{tcolorbox}

\paragraph{PrimeIntellect.} The user prompt is the dataset's \verb|problem| field as-is:

\begin{tcolorbox}[enhanced, breakable, title={User Prompt (PrimeIntellect)}, colback=user-bg, colframe=user-frame, colbacktitle=user-frame!30!white, coltitle=black, fonttitle=\bfseries, boxrule=0.5mm, arc=2mm]
\begin{verbatim}
{{problem}}
\end{verbatim}
\end{tcolorbox}

\paragraph{TACO.} The TACO prompt template is:

\begin{tcolorbox}[enhanced, breakable, title={User Prompt (TACO, base template)}, colback=user-bg, colframe=user-frame, colbacktitle=user-frame!30!white, coltitle=black, fonttitle=\bfseries, boxrule=0.5mm, arc=2mm]
\begin{verbatim}
Solve the programming task below in a Python markdown code block.

{{question}}

{{starter_code_instruction_or_stdin_instruction}}
\end{verbatim}
\end{tcolorbox}

The placeholder \verb|{{starter_code_instruction_or_stdin_instruction}}| is replaced by one of the two snippets below depending on whether the problem provides starter code or expects STDIN-based I/O:

\begin{tcolorbox}[enhanced, breakable, title={TACO instruction: with starter code}, colback=user-bg, colframe=user-frame, colbacktitle=user-frame!30!white, coltitle=black, fonttitle=\bfseries, boxrule=0.5mm, arc=2mm]
\begin{verbatim}
You will use the following starter code to write the solution to the
problem and enclose your code within ```python delimiters'''.
```python
{{starter_code}}
'''
\end{verbatim}
\end{tcolorbox}

\begin{tcolorbox}[enhanced, breakable, title={TACO instruction: STDIN}, colback=user-bg, colframe=user-frame, colbacktitle=user-frame!30!white, coltitle=black, fonttitle=\bfseries, boxrule=0.5mm, arc=2mm]
\begin{verbatim}
Read the inputs from stdin, solve the problem, and write the answer to
stdout (do not directly test on the sample inputs). Enclose your code
within ```python delimiters'''.
\end{verbatim}
\end{tcolorbox}

\subsection{Logic and Visual Reasoning}

\begin{tcolorbox}[enhanced, breakable, title={User Prompt (ARC-AGI and BARC)}, colback=user-bg, colframe=user-frame, colbacktitle=user-frame!30!white, coltitle=black, fonttitle=\bfseries, boxrule=0.5mm, arc=2mm]
\begin{verbatim}
You are a world-class puzzle solver with exceptional pattern
recognition skills. Your task is to analyze puzzles, spot patterns,
and provide direct solutions.

Given input-output grid pairs as reference examples, carefully observe
the patterns to predict the output grid for new test input. Each pair
follows the same transformation rule. Grids are 2D arrays. Here are
the input and output grids for the reference examples:
----------------------------------------
{{training_data}}
----------------------------------------
Now, solve the following puzzle based on its input grid by applying
the rules you have learned from the training data.
----------------------------------------
{{input_test_data}}
----------------------------------------
What is the output grid? Please put your answer within <answer> and
</answer> tags, your final answer should be only the output grid
(2d array).
\end{verbatim}
\end{tcolorbox}

\begin{tcolorbox}[enhanced, breakable, title={User Prompt (Graph Logical Reasoning)}, colback=user-bg, colframe=user-frame, colbacktitle=user-frame!30!white, coltitle=black, fonttitle=\bfseries, boxrule=0.5mm, arc=2mm]
\begin{verbatim}
{{question}} Please put your answer within <answer> and </answer>
tags, for example <answer> fdebme </answer>.
\end{verbatim}
\end{tcolorbox}

\begin{tcolorbox}[enhanced, breakable, title={User Prompt (Ordering Puzzle)}, colback=user-bg, colframe=user-frame, colbacktitle=user-frame!30!white, coltitle=black, fonttitle=\bfseries, boxrule=0.5mm, arc=2mm]
\begin{verbatim}
{{instruction}}. The constraints are: {{constraints}}. Please put
your answer within <answer> and </answer> tags, for example
<answer> [`pigeon', `sparrow', `quail'] </answer>.
\end{verbatim}
\end{tcolorbox}

\begin{tcolorbox}[enhanced, breakable, title={User Prompt (Zebra Puzzle)}, colback=user-bg, colframe=user-frame, colbacktitle=user-frame!30!white, coltitle=black, fonttitle=\bfseries, boxrule=0.5mm, arc=2mm]
\begin{verbatim}
{{instruction}} The clues are: {{clues}}. Output the grid in the form
of a dictionary with keys as header containing a list of the attributes
and rows denoting each row of the final grid. Please return the final
answer in <answer> </answer> tags, for example <answer>
{"header": ["Position", "Nationality", "Job"],
 "rows": [["1", "british", "plumber"],
          ["2", "polish", "carpenter"]]} </answer>.
\end{verbatim}
\end{tcolorbox}

\subsection{Simulation}

\begin{tcolorbox}[enhanced, breakable, title={User Prompt (CodeI/O, Input Prediction)}, colback=user-bg, colframe=user-frame, colbacktitle=user-frame!30!white, coltitle=black, fonttitle=\bfseries, boxrule=0.5mm, arc=2mm]
\begin{verbatim}
You are given a question that requires some input and output variables
as follows:

{{problem_description}}

The input and output requirements are as follows:

{{io_requirements}}

Given the following output:

{{given}}

Can you predict a feasible input without writing any code? Please
reason and put your final answer in the following json format:
"input": <your input>, where <your input> should be a dictionary, even
if there is only one input variable, with keys strictly matching the
input variables' names as specified. Please put your answer in
\boxed{} tags.

Tip: Here is a reference code snippet for this question. You can refer
to this code to guide your reasoning but not copy spans of code
directly.

{{refcode}}

Please output the final answer in JSON format.
\end{verbatim}
\end{tcolorbox}

\begin{tcolorbox}[enhanced, breakable, title={User Prompt (CodeI/O, Output Prediction)}, colback=user-bg, colframe=user-frame, colbacktitle=user-frame!30!white, coltitle=black, fonttitle=\bfseries, boxrule=0.5mm, arc=2mm]
\begin{verbatim}
You are given a question that requires some input and output variables
as follows:

{{problem_description}}

The input and output requirements are as follows:

{{io_requirements}}

Given the following input:

{{given}}

Can you predict the output without writing any code? Please reason and
put your final answer in the following json format:
"output": <your output>, where <your output> should strictly match the
output requirement as specified. Please put your answer in
\boxed{} tags.

Tip: Here is a reference code snippet for this question. You can refer
to this code to guide your reasoning but not copy spans of code
directly.

{{refcode}}

Please output the final answer in JSON format.
\end{verbatim}
\end{tcolorbox}

\subsection{Table Reasoning}

\begin{tcolorbox}[enhanced, breakable, title={User Prompt (HiTab)}, colback=user-bg, colframe=user-frame, colbacktitle=user-frame!30!white, coltitle=black, fonttitle=\bfseries, boxrule=0.5mm, arc=2mm]
\begin{verbatim}
You are given one or more tables. Use the information in the table to
answer the following question.

{{tables}}

The question is:

{{question}}

Please output the final answer within \boxed{}. If there are multiple
answers, please output them separated by |.
\end{verbatim}
\end{tcolorbox}

\begin{tcolorbox}[enhanced, breakable, title={User Prompt (MultiHierTT)}, colback=user-bg, colframe=user-frame, colbacktitle=user-frame!30!white, coltitle=black, fonttitle=\bfseries, boxrule=0.5mm, arc=2mm]
\begin{verbatim}
You are given one or more tables. Use the information in the tables to
answer the following question.

{{tables}}

The question is:

{{question}}

Please output the final answer within \boxed{}.
\end{verbatim}
\end{tcolorbox}

\subsection{STEM}

For \textsc{OpenScienceReasoning-2}, the user prompt is the source dataset's \verb|input| field as-is, with no additional template wrapping.

\section{Implementation Details}
\label{app:im_details}

\subsection{Datasets}
\label{app:datasets}

\paragraph{Training data.}
We follow the \textsc{GURU} multi-domain reasoning suite~\citep{chengrevisiting} and aggregate training data from six high-level domains, replacing only the \textit{stem} source. \textsc{GURU} provides per-domain corpora that have already been deduplicated, heuristically filtered, and difficulty-filtered using a weak/strong model pass-rate scheme (see~\citet{chengrevisiting} for the full pipeline). We use these curated subsets directly, without re-applying any of \textsc{GURU}'s upstream filtering steps. Per-domain source files and raw sizes after \textsc{GURU}'s curation are summarized in Table~\ref{tab:train_sources}.
\vspace{-5pt}
\paragraph{Per-domain sources.} We use the following sources, all inherited from \textsc{GURU}~\citep{chengrevisiting} except where noted:
\begin{itemize}[leftmargin=1.2em, itemsep=1mm, topsep=1pt]
    \item \textit{Math.} Aggregated from OR1~\citep{he2025skywork}, DAPO~\citep{yu2025dapo}, and DeepScaler~\citep{luo2025deepscaler}, which themselves compile competition-style problems including AIME~\citep{AoPS_AIME} and AMC~\citep{maa_amc}.
    \item \textit{Codegen.} LeetCode~\citep{xia2025leetcodedataset}, TACO-Verified~\citep{li2024taco}, PrimeIntellect~\citep{mattern2025synthetic}, and LiveCodeBench~\citep{jain2024livecodebench}, with the PrimeIntellect and LiveCodeBench subsets adopted from the pre-filtered DeepCoder release~\citep{luo2025deepcoder}.
    \item \textit{Logic.} ARC-AGI-1/2~\citep{chollet2024arc, chollet2025arc2} and BARC~\citep{li2024combining} (existing datasets), together with three synthesized symbolic-reasoning tasks introduced by \textsc{GURU}: Zebra Puzzle (following~\citet{lin2025zebralogic}), Ordering Puzzle, and Graph Puzzle (following~\citet{saparov2023language, saparov2025transformers}).
    \item \textit{Simulation.} Code I/O on PyEdu~\citep{li2025codeio}, where the model predicts program inputs from outputs or outputs from inputs without executing code.
    \item \textit{Table.} HiTab~\citep{cheng2022hitab} and MultiHierTT~\citep{zhao2022multihiertt}, both linearized into markdown format. When multiple tables accompany a single query, they are concatenated with line breaks.
    \item \textit{Stem.} We depart from \textsc{GURU}'s default WebInstruct-Verified~\citep{ma2025generalreasoner} source and instead use NVIDIA's OpenScienceReasoning-2\footnote{\url{https://huggingface.co/datasets/nvidia/OpenScienceReasoning-2}}, which provides higher-quality reasoning traces and is formatted as multiple-choice questions with ground-truth answers. The latter point lets us replace \textsc{GURU}'s 1.5B model-based verifier~\citep{ma2025generalreasoner} with simple rule-based answer matching, eliminating verifier-model noise from the stem reward signal.
\end{itemize}
\vspace{-5pt}
\paragraph{Reward design.} Following \textsc{GURU}, all domains use binary verifiable rewards (1 if correct, 0 otherwise). With our stem replacement, the entire training suite uses only two verification modes: \textit{(i)} rule-based matching after extracting answers from \verb|\boxed{}| or \verb|<answer>| tags, used for math, logic, simulation, table, and stem (the last via multiple-choice answer extraction); and \textit{(ii)} execution-based verification, where generated programs are run against test cases under a 30-second timeout and 10\,GB memory limit, with reward 1 granted only if all test cases pass (codegen). Notably, our pipeline avoids \textsc{GURU}'s third reward mode---model-based verification with a 1.5B verifier for stem---which removes a learned-component noise source from RL training.
\begin{table}[t]
\centering
\small
\caption{Training data sources by domain. Sizes are GURU's post-curation per-source counts (i.e., \emph{before} our subsampling). Codegen and logic each pool multiple sub-sources; STEM uses our replacement source.}
\label{tab:train_sources}
\begin{tabular}{lll}
\toprule
\textbf{Domain} & \textbf{Source} & \textbf{Pre-subsample size} \\
\midrule
Math & OR1 / DAPO / DeepScaler~\citep{he2025skywork, yu2025dapo, luo2025deepscaler} & 54.4k \\
\midrule
\multirow{4}{*}{Codegen} & LeetCode~\citep{xia2025leetcodedataset} & 1.3k \\
                         & LiveCodeBench~\citep{jain2024livecodebench} & 0.4k \\
                         & PrimeIntellect~\citep{mattern2025synthetic} & 7.5k \\
                         & TACO-Verified~\citep{li2024taco} & 8.8k \\
\midrule
\multirow{6}{*}{Logic} & ARC-AGI-1~\citep{chollet2024arc} & 0.1k \\
                       & ARC-AGI-2~\citep{chollet2025arc2} & 0.2k \\
                       & BARC~\citep{li2024combining} & 1.6k \\
                       & Graph Puzzle~\citep{saparov2025transformers} & 1.2k \\
                       & Ordering Puzzle & 1.9k \\
                       & Zebra Puzzle~\citep{lin2025zebralogic} & 1.3k \\
\midrule
Simulation & Code I/O (PyEdu)~\citep{li2025codeio} & 3.7k \\
\midrule
\multirow{2}{*}{Table} & HiTab~\citep{cheng2022hitab} & 4.3k \\
                       & MultiHierTT~\citep{zhao2022multihiertt} & 1.5k \\
\midrule
STEM & OpenScienceReasoning-2 & 1.4M \\
\bottomrule
\end{tabular}
\end{table}
\vspace{-5pt}
\paragraph{Subsampling.}
The raw mixture is dominated by math and stem by two orders of magnitude (Table~\ref{tab:train_sources}), which would obscure the contribution of any curriculum if used directly. We therefore subsample the training set in two stages, governed by a fixed seed (\texttt{subsample\_seed}=42) so every method trains on exactly the same data. First, each source file is uniformly subsampled at ratio 0.2. Second, the per-source results are pooled by domain and capped at a per-domain budget; within a domain, the cap is applied evenly across sub-sources (e.g., codegen draws roughly evenly from its four sources). The two training mixtures we use differ only in this per-domain cap, summarized in Table~\ref{tab:train_caps}.
\begin{table}[t]
\centering
\small
\caption{Per-domain training caps after subsampling. The balanced mixture is used for the main results (Table~\ref{tab:main_results}); the imbalanced mixture is used for the stress test (Table~\ref{tab:imbalanced_results}).}
\label{tab:train_caps}
\begin{tabular}{lcccccc}
\toprule
\textbf{Mixture} & \textbf{Math} & \textbf{Codegen} & \textbf{Logic} & \textbf{Simulation} & \textbf{Table} & \textbf{STEM} \\
\midrule
Balanced (main) & 1{,}000 & 1{,}000 & 1{,}000 & 1{,}000 & 1{,}000 & 1{,}000 \\
Imbalanced      & 1{,}500 & 1{,}000 & 1{,}000 & \phantom{1{,}}500 & \phantom{1{,}}500 & 1{,}500 \\
\bottomrule
\end{tabular}
\end{table}
\vspace{-5pt}
\paragraph{Validation set (online evaluation).}
During training we periodically evaluate the policy on a validation set drawn from the same six domains, used \emph{only} for monitoring training dynamics. We do not perform validation-based checkpoint selection or hyperparameter tuning---all numbers reported in Table~\ref{tab:main_results} come from each run's \emph{final} checkpoint (see Appendix~\ref{app:eval_details}). This validation set partially overlaps with the held-out evaluation benchmarks reported in Table~\ref{tab:main_results}; because no model-selection or tuning decision is made on the basis of validation accuracy, this overlap does not affect the held-out numbers. Per-domain validation subsets:
\begin{itemize}[leftmargin=1.2em, itemsep=1mm, topsep=1pt]
    \item \textit{Math.} MATH-500 (500 samples), AIME (240 samples, 8$\times$ repetition), AMC (332 samples, 4$\times$ repetition).
    \item \textit{Codegen.} HumanEval (164 samples), MBPP (200 samples), LiveCodeBench (279 samples).
    \item \textit{Logic.} Ordering puzzles (100 samples), Zebra puzzles (200 samples), ARC-AGI-1 (200 samples).
    \item \textit{Simulation.} CodeI/O (200 samples).
    \item \textit{Table.} HiTab (200 samples), MultiHierTT (200 samples).
    \item \textit{Stem.} SuperGPQA (200 samples).
\end{itemize}
\vspace{-5pt}
\paragraph{Evaluation set.}
The held-out evaluation benchmarks reported in the main paper (Table~\ref{tab:main_results}) cover the same 14 datasets across all six domains. To reduce evaluation noise, each benchmark is run multiple times per checkpoint, with the per-benchmark run multipliers shown in Table~\ref{tab:eval_runs}. The reported numbers in Table~\ref{tab:main_results} are means over three training seeds with each entry additionally averaged over the corresponding number of evaluation runs (so most cells reflect $3 \times \{1,4,32\}$ samples per benchmark). All reported numbers come from each run's \emph{final} checkpoint; we do not select checkpoints by validation accuracy.

\begin{table}[t]
\centering
\small
\caption{Evaluation run multipliers per benchmark. Each benchmark is evaluated this many times per checkpoint, and the per-checkpoint result is the mean across runs. AIME applies an additional 8$\times$ in-set repetition before the run multiplier (32 effective samples per problem).}
\label{tab:eval_runs}
\begin{tabular}{lc}
\toprule
\textbf{Benchmark} & \textbf{Runs per checkpoint} \\
\midrule
MATH-500, MBPP, HiTab, SuperGPQA & 1 \\
HumanEval, GPQA-Diamond, MultiHierTT, CruxEval & 4 \\
CodeI/O, ARC-AGI-1, Zebra Puzzle, FinQA & 4 \\
AIME & 32 \\
\bottomrule
\end{tabular}
\end{table}

\subsection{Training Details}
\label{app:training_details}

We implement \alg on top of the verl framework using the DAPO recipe
(\texttt{recipe.dapo.main\_dapo}), which provides a GRPO-style policy update with
asymmetric clipping and dynamic batching. All baselines (Random, Math-to-Others,
SEC) share the same DAPO configuration; only the data-sampling component differs.
We split the configuration into method-related settings, which control the
\alg curriculum, and RL-related settings, which control the underlying
GRPO/DAPO update. 
\vspace{-5pt}
\paragraph{Method-related settings (\alg).}
The bandit operates on the six domains extracted from each example's \texttt{data\_source} field. $Q$-values are initialized centered log-proportional to domain size (Eq.~\ref{eq:q_init}, $\kappa{=}0.5$), and five rounds ($W{=}5$) of round-robin warmup per epoch precede bandit control. When a domain's index pool is exhausted within an epoch, we reset that pool and continue sampling from it, so high-$Q$ domains can be over-sampled within an epoch while no data is permanently withheld across epochs. The transferability signal is a Rademacher JL projection~\citep{park2023trak} of the per-domain gradient computed on the last $N{=}4$ transformer layers, with a fixed projection seed shared across all runs. The full hyperparameter list is given in Table~\ref{tab:method_hp}.
\begin{table}[t]
\centering
\small
\caption{\alg method-related hyperparameters.}
\label{tab:method_hp}
\begin{tabular}{lll}
\toprule
\textbf{Component} & \textbf{Hyperparameter} & \textbf{Value} \\
\midrule
\multirow{7}{*}{Bandit}
 & Selection rule                  & Boltzmann sampling over UCB-augmented $Q$ (Eq.~\ref{eq:sampling}) \\
 & Softmax temperature $\tau$      & $0.85$ \\
 & EMA learning rate $\alpha$      & $0.3$ \\
 & UCB exploration coefficient $c$ & $0.2$ \\
 & Warmup rounds (round-robin) $W$ & $5$ per epoch \\
 & Initial $Q$ distribution        & centered log-proportional to domain size (Eq.~\ref{eq:q_init}, $\kappa{=}0.5$) \\
 & Exhaustion strategy             & reset \\
\midrule
\multirow{4}{*}{Composite signal}
 & Mixing coefficient $\beta$      & $0.2$ \\
 & Learnability normalization      & EMA z-score (Eq.~\ref{eq:learnability_norm}), decay $0.8$, warmup $12$ steps \\
 & Comparison frequency $K_c$      & $2$ steps \\
 & Value update                    & two-phase (sampled arm every step; all arms at refresh steps) \\
\midrule
\multirow{9}{*}{Transferability}
 & Type                            & projected-gradient cosine (Eqs.~\ref{eq:raw_transfer},~\ref{eq:minmax_transfer}) \\
 & Parameter subset                & last $N{=}4$ transformer layers \\
 & Projection source               & final PPO mini-batch of each step \\
 & Projection method               & Rademacher JL~\citep{park2023trak} \\
 & Projection dimension $r$        & $4096$ \\
 & Per-domain gradient EMA $\gamma$ & $0.8$ \\
 & Temporal smoothing $\delta$     & $0.8$ \\
 & Cross-domain normalization      & min--max (Eq.~\ref{eq:minmax_transfer}), scale-EMA $\delta_s{=}0.9$ \\
 & Scale floor $s_{\min}$           & $0.01$ \\
\bottomrule
\end{tabular}
\end{table}
\vspace{-5pt}
\paragraph{Two-phase update bookkeeping.}
Because transferability is recomputed for every domain at each comparison step, \alg updates the unsampled arms there as well as the sampled one (\S\ref{subsec:tac_rl}). Two details govern how the learnability term enters those updates. First, what is stored per arm is the \emph{normalized} learnability $\hat{L}_m$ (Eq.~\ref{eq:learnability_norm}) from the last step the arm was pulled, not the raw $L_m$. When an unsampled arm is updated, this cached $\hat{L}_m$ is reused as-is and is \emph{not} re-passed through the current normalizer. Re-normalizing a stale raw value through the live running mean and standard deviation, which keep drifting as other arms are sampled, would let an arm's effective learnability change without any new rollout for it; caching the already-normalized value pins each arm's learnability to the last time it was actually measured. Second, the normalizer statistics $\mu_L^{(t)}, \sigma_L^{(t)}$ are updated only on the sampled arm's fresh $L_{m_t}^{(t)}$, never during the unsampled-arm updates, so the running statistics track only genuinely observed advantages. Arms never yet pulled have no cached $\hat{L}_m$ and are skipped entirely until their first selection.
\vspace{-5pt}
\paragraph{RL-related settings (DAPO/GRPO).}
We use a DAPO-style~\citep{yu2025dapo} update with grouped advantages, asymmetric clipping ($\epsilon_\text{low}{=}0.2$, $\epsilon_\text{high}{=}0.4$), no KL regularization, and dynamic per-GPU token budgets. The base models are \texttt{Qwen3-1.7B-Base}~\citep{yang2025qwen3} and \texttt{Llama3.2-3B-Instruct}~\citep{grattafiori2024llama}; both are trained with FSDP~\citep{zhao2023pytorch} and full parameter/optimizer offloading. Generation uses vLLM~\citep{kwon2023efficient} in synchronous mode. Validation evaluation is run before the first training step and every 20 update steps thereafter. The detailed configuration is given in Table~\ref{tab:rl_hp}.
\begin{table}[t]
\centering
\small
\caption{RL-related (DAPO/GRPO) hyperparameters.}
\label{tab:rl_hp}
\begin{tabular}{lll}
\toprule
\textbf{Component} & \textbf{Hyperparameter} & \textbf{Value} \\
\midrule
\multirow{4}{*}{Algorithm}
 & Advantage estimator             & GRPO (group-relative) \\
 & Clip ratio (low / high)         & $0.2$ / $0.4$ \\
 & Use KL in reward / loss         & False / False \\
 & Loss aggregation                & token-mean \\
\midrule
\multirow{5}{*}{Optimizer}
 & Optimizer                       & AdamW \\
 & Learning rate                   & $1\times 10^{-6}$ \\
 & LR schedule / warmup            & constant, $10$ steps \\
 & Weight decay                    & $0.1$ \\
 & Gradient clipping               & $1.0$ \\
\midrule
\multirow{4}{*}{Batching}
 & Train prompt batch size         & $64$ (single-domain) \\
 & PPO mini-batch size             & $16$ \\
 & Rollouts per prompt $K$         & $4$ \\
 & Max tokens / GPU                & $2(L_p{+}L_r) = 24{,}576$ \\
\midrule
\multirow{4}{*}{Sequence}
 & Max prompt length $L_p$         & $4096$ \\
 & Max response length $L_r$       & $8192$ \\
 & Sequence parallelism            & $1$ \\
 & Tensor parallelism (rollout)    & $1$ \\
\midrule
\multirow{4}{*}{Sampling}
 & Temperature                     & $1.0$ \\
 & top-$p$                         & $1.0$ \\
 & vLLM GPU memory utilisation     & $0.7$ \\
\midrule
\multirow{1}{*}{Trainer}
 & Total epochs                    & $2$ \\
\bottomrule
\end{tabular}
\end{table}
\vspace{-5pt}
\paragraph{Compute.}
All training runs use a single node with 4$\times$ NVIDIA H100 80\,GB GPUs, 16 CPU cores, and 512\,GB of system memory. With FSDP, full parameter/optimizer offloading, and a 12K-token context window, two epochs of GRPO on the balanced mixture take approximately 7 hours of wall-clock time per seed (${\approx}28$ GPU-hours). \alg adds less than $1\%$ wall-clock overhead per training step relative to the other curricula (Figure~\ref{fig:walltime}), so the same compute budget is used for every method to ensure a fair comparison.

\subsection{Evaluation Details}
\label{app:eval_details}
\paragraph{Pipeline.}
Held-out evaluation is run offline after training completes, on saved checkpoints. For each checkpoint, we merge the trained model from its sharded format into a HuggingFace-format model, load it into vLLM, generate responses for every benchmark in a single inference session to amortize the engine load cost, and score the outputs against verifiable references on CPU. Generation outputs are cached to disk, allowing the scoring pass to be re-run without regenerating responses if scoring logic changes.
\vspace{-5pt}
\paragraph{Generation settings.}
The vLLM engine is initialized once per checkpoint with the maximum prompt and response lengths across all benchmarks, so a single engine serves every benchmark without reloading weights. Sampling uses temperature $1.0$ and top-$p$ $1.0$, matching the training-time sampling configuration. Engine-wide settings are listed in Table~\ref{tab:rl_hp}.
\vspace{-5pt}
\paragraph{Scoring and aggregation.}
Each benchmark uses a verifiable reward function appropriate to its domain: exact-match against ground truth for math (after \verb|\boxed{}| extraction), unit-test execution for codegen, structural answer parsing for logic and table benchmarks, and multiple-choice answer extraction for stem. Each benchmark produces a single mean accuracy. We aggregate within each domain by taking the unweighted mean of its benchmark accuracies, then report the unweighted mean across the six domains as the overall score in Table~\ref{tab:main_results}. All numbers are means over three training seeds; the seeds vary the bandit sampling sequence, while the data subsampling seed is held fixed across methods so every method trains on identical data.

\section{More Experiments}
\label{app:more_exps}

\begin{figure}[t]
\centering
\begin{subfigure}[t]{0.34\textwidth}
    \centering
    \includegraphics[width=\textwidth]{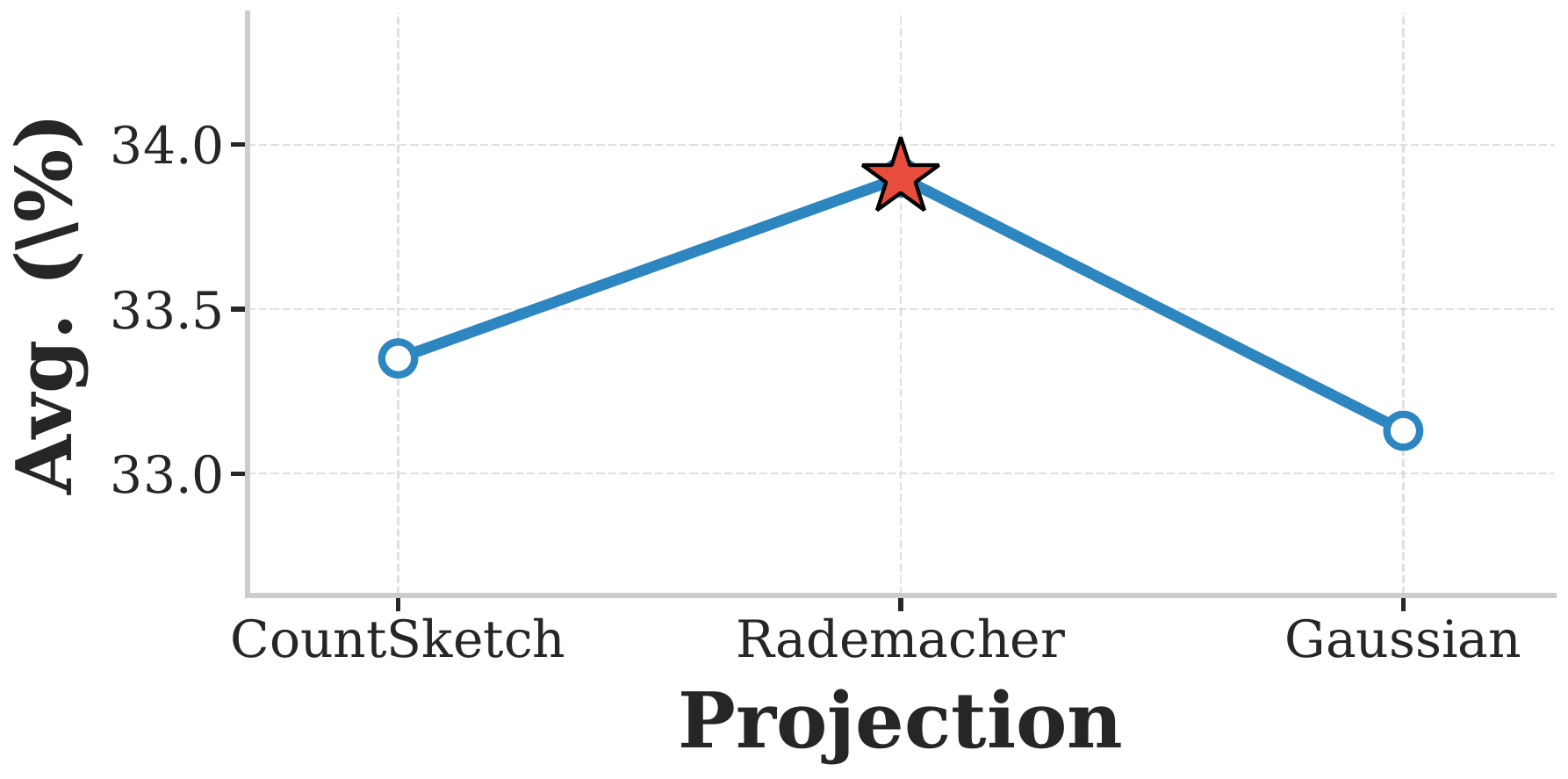}
    \caption{Projection method.}
    \label{fig:ablation_projection}
\end{subfigure}
\hfill
\begin{subfigure}[t]{0.34\textwidth}
    \centering
    \includegraphics[width=\textwidth]{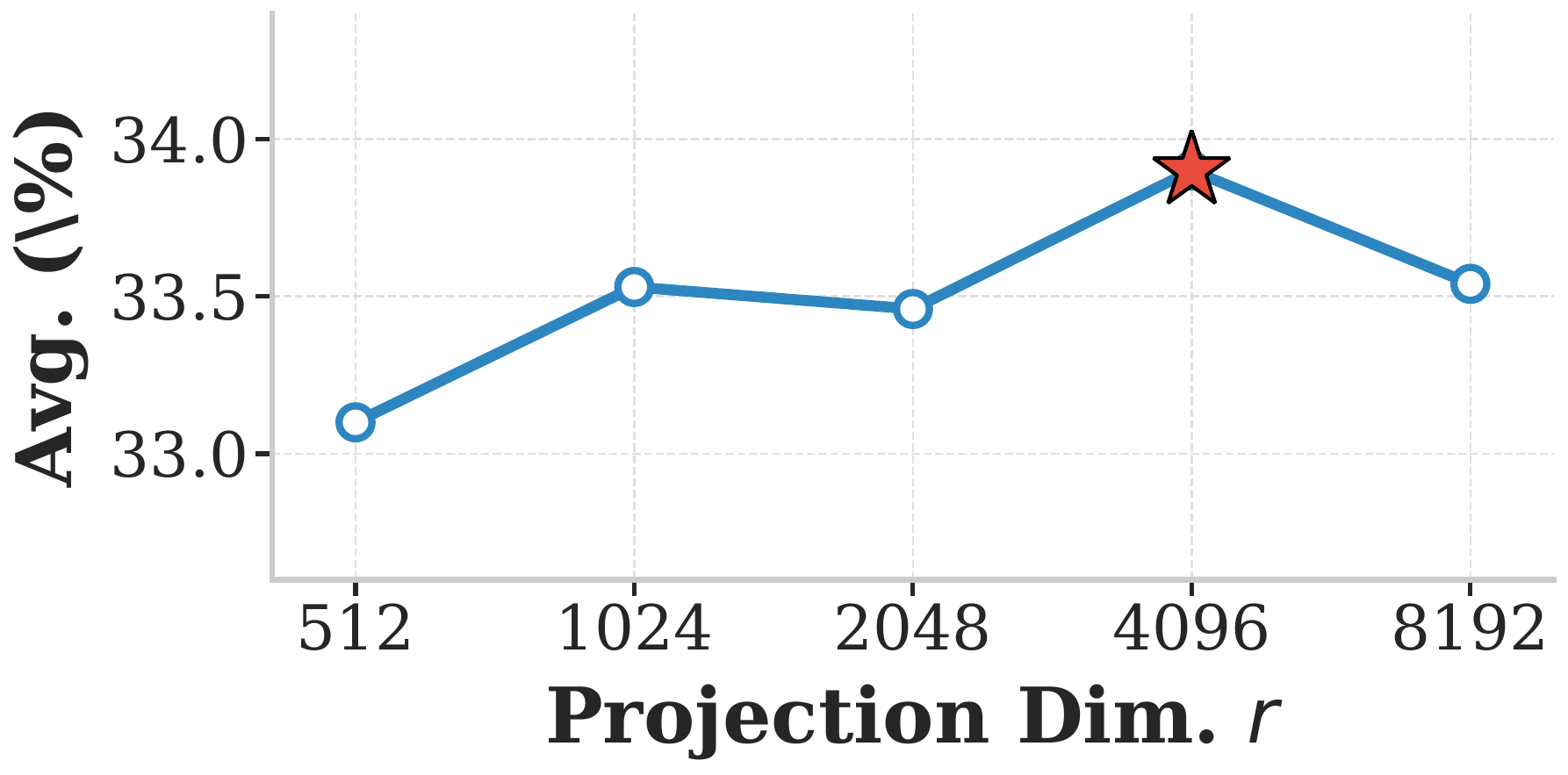}
    \caption{Projection dimension $r$.}
    \label{fig:ablation_projdim}
\end{subfigure}
\hfill
\begin{subfigure}[t]{0.29\textwidth}
    \centering
    \includegraphics[width=\textwidth]{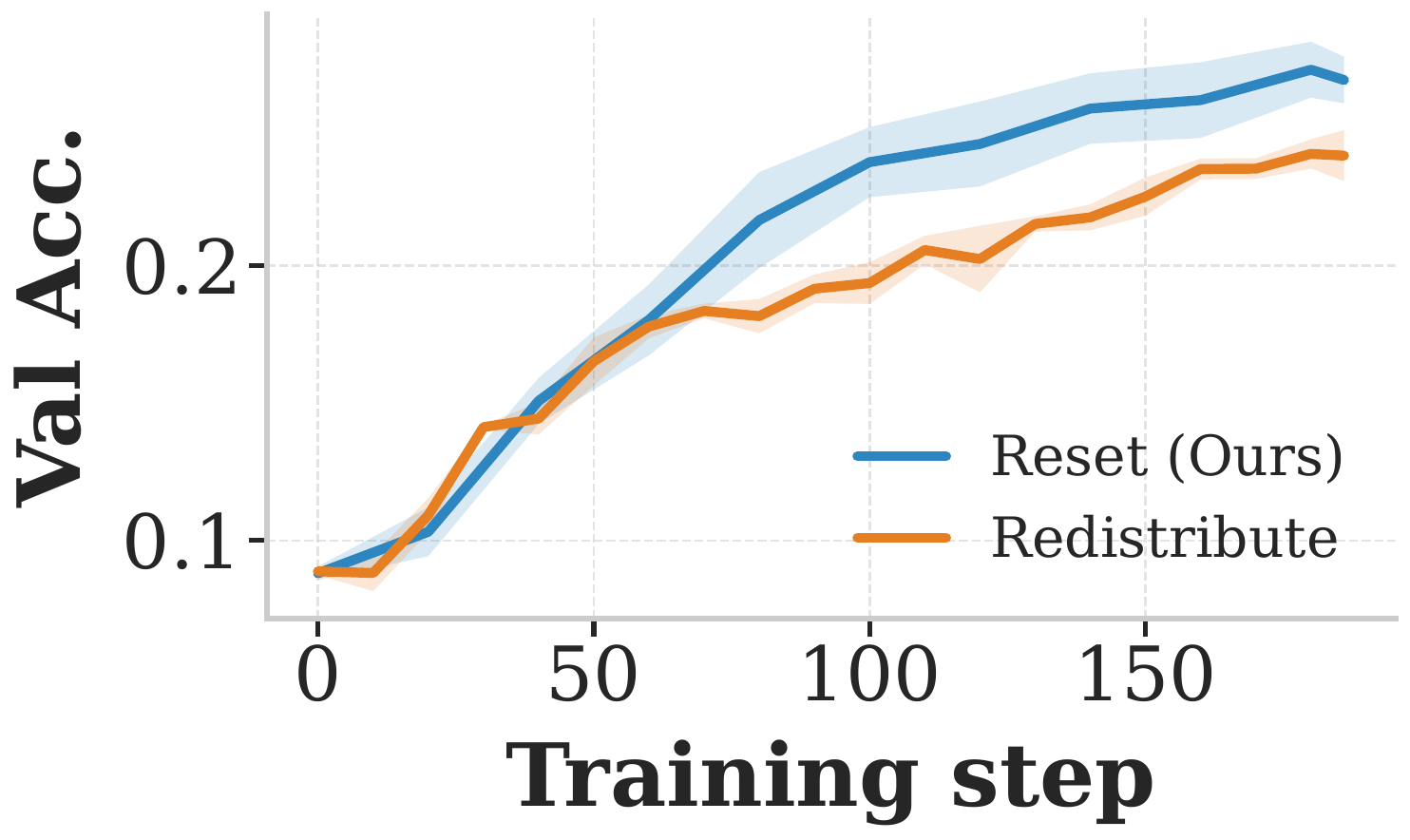}
    \caption{Data-exhaustion strategy.}
    \label{fig:ablation_exhaustion}
\end{subfigure}
\caption{\textbf{Additional ablations.} \textbf{(a)} Macro-averaged accuracy (unweighted mean over the six domain scores) across choices of random projection family (CountSketch, Rademacher, Gaussian). \textbf{(b)} Macro accuracy across projection dimensions $r \in \{512, 1024, 2048, 4096, 8192\}$. \textbf{(c)} Validation accuracy over training under two data-exhaustion strategies: \emph{reset} (\alg's default) reshuffles a depleted domain in place; \emph{redistribute} re-draws from the remaining domains.}
\label{fig:additional_ablations}
\vspace{-5pt}
\end{figure}
\subsection{More Ablation Study}
\label{app:more_ablations}
We complement the main hyperparameter sweeps in Section~\ref{subsec:results}~(Figure~\ref{fig:ablations}) with three additional ablations targeting design choices in the transferability pipeline. Figure~\ref{fig:additional_ablations} reports macro-averaged accuracy across all 14 evaluation benchmarks; we vary one factor at a time while holding all other hyperparameters at their default values from Table~\ref{tab:method_hp}.
\vspace{-5pt}
\paragraph{Projection method.}
We compare the three random projection families used in the gradient-attribution literature: CountSketch~\citep{charikar2002finding}, Rademacher~\citep{achlioptas2003database} (\alg's default), and Gaussian. As shown in Figure~\ref{fig:ablation_projection}, the Rademacher projection outperforms the others by roughly $0.5$--$0.7$ points, while CountSketch and Gaussian are within $0.2$ points of each other. We attribute Rademacher's edge to its sub-Gaussian tail behavior, which preserves inner products with low variance at the projection dimension ($r{=}4096$) used here, consistent with its use as the default in TRAK~\citep{park2023trak}.
\vspace{-5pt}
\paragraph{Projection dimension.}
We sweep the projection dimension $r \in \{512, 1024, 2048, 4096, 8192\}$. Performance improves up to $r{=}4096$ and then flattens, with $r{=}8192$ giving no further gain and the smaller dimensions losing roughly a point of macro accuracy (Figure~\ref{fig:ablation_projdim}). The plateau at the top end is consistent with the Johnson-Lindenstrauss bound~\citep{johnson1984extensions}: once $r$ is large enough to preserve pairwise inner products at the granularity the cosine signal needs, further increases yield diminishing returns. We use $r{=}4096$ as the default, at the knee of the curve.
\vspace{-5pt}
\paragraph{Data-exhaustion strategy.}
When a domain's index pool is exhausted within an epoch, the curriculum must decide what to do at the next time the bandit selects that domain. We compare two strategies: \emph{reset} (\alg's default), which reshuffles the depleted domain's data and continues sampling from it; and \emph{redistribute}, which re-draws $m_t$ from the remaining domains. Figure~\ref{fig:ablation_exhaustion} shows that \emph{reset} converges to higher validation accuracy throughout training, with the gap widening as more domains exhaust. \emph{Redistribute} forces the sampler to pick among lower-$Q$ domains late in an epoch simply because their data happens to remain, dragging down the effective curriculum quality. Resetting instead lets high-$Q$ domains continue to be over-sampled within an epoch, while still guaranteeing that no data is permanently withheld across epochs.

\subsection{Complexity Analysis}
\label{app:complexity}
\begin{figure}[t]
\centering
\begin{subfigure}[t]{0.32\textwidth}
    \centering
    \includegraphics[width=\textwidth]{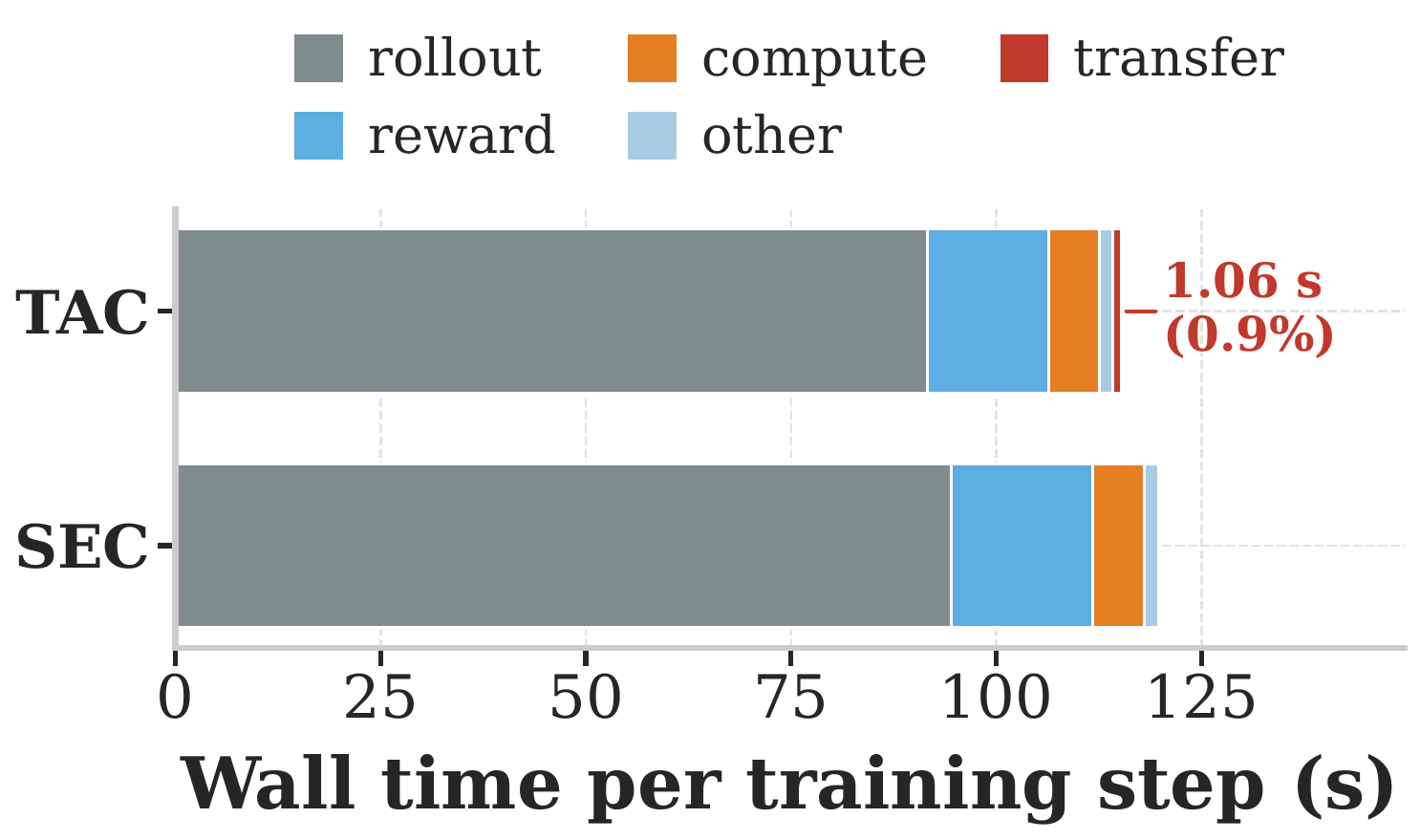}
    \caption{Per-step time breakdown.}
    \label{fig:walltime_breakdown}
\end{subfigure}
\hfill
\begin{subfigure}[t]{0.32\textwidth}
    \centering
    \includegraphics[width=\textwidth]{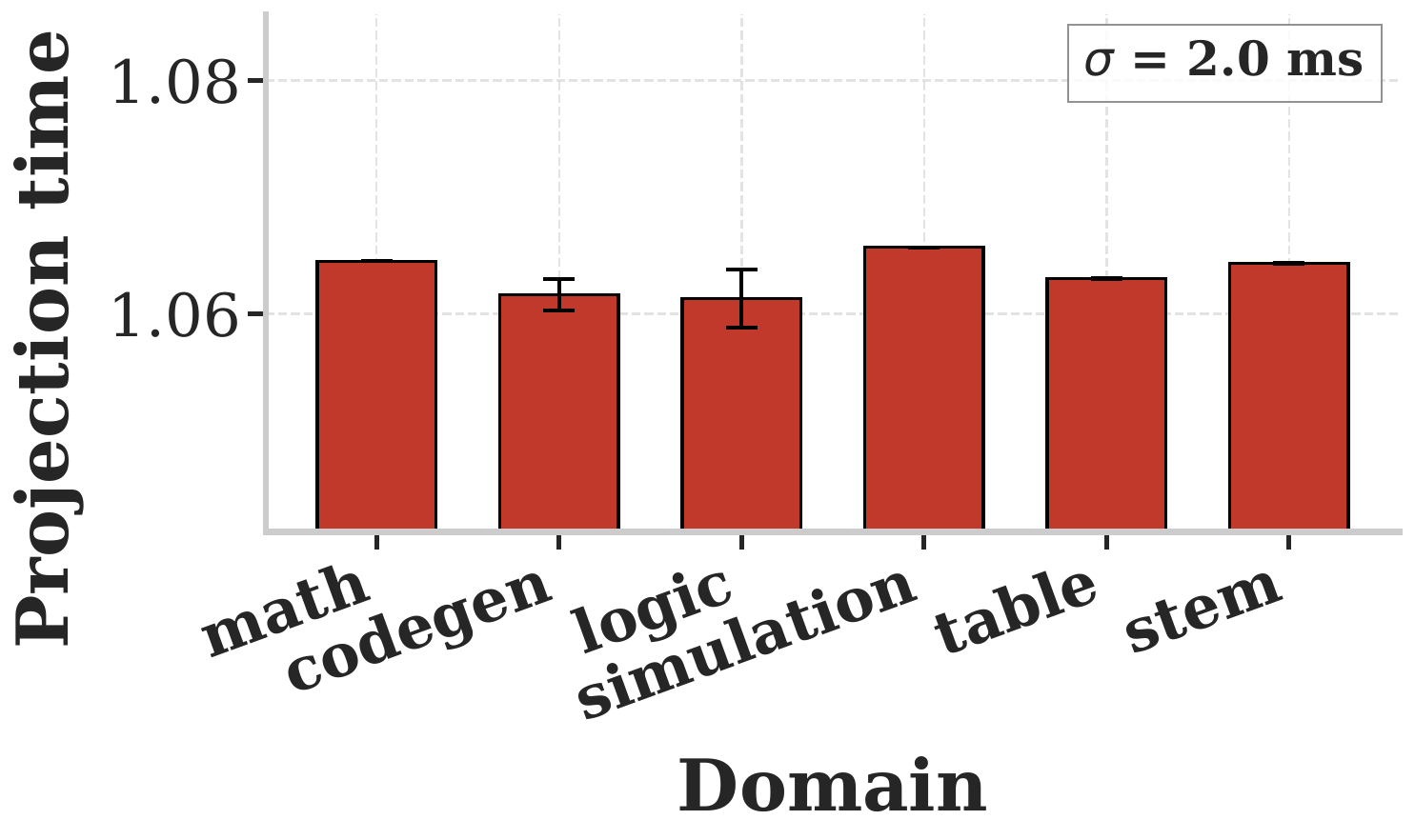}
    \caption{Projection cost by domain.}
    \label{fig:walltime_per_domain}
\end{subfigure}
\hfill
\begin{subfigure}[t]{0.32\textwidth}
    \centering
    \includegraphics[width=\textwidth]{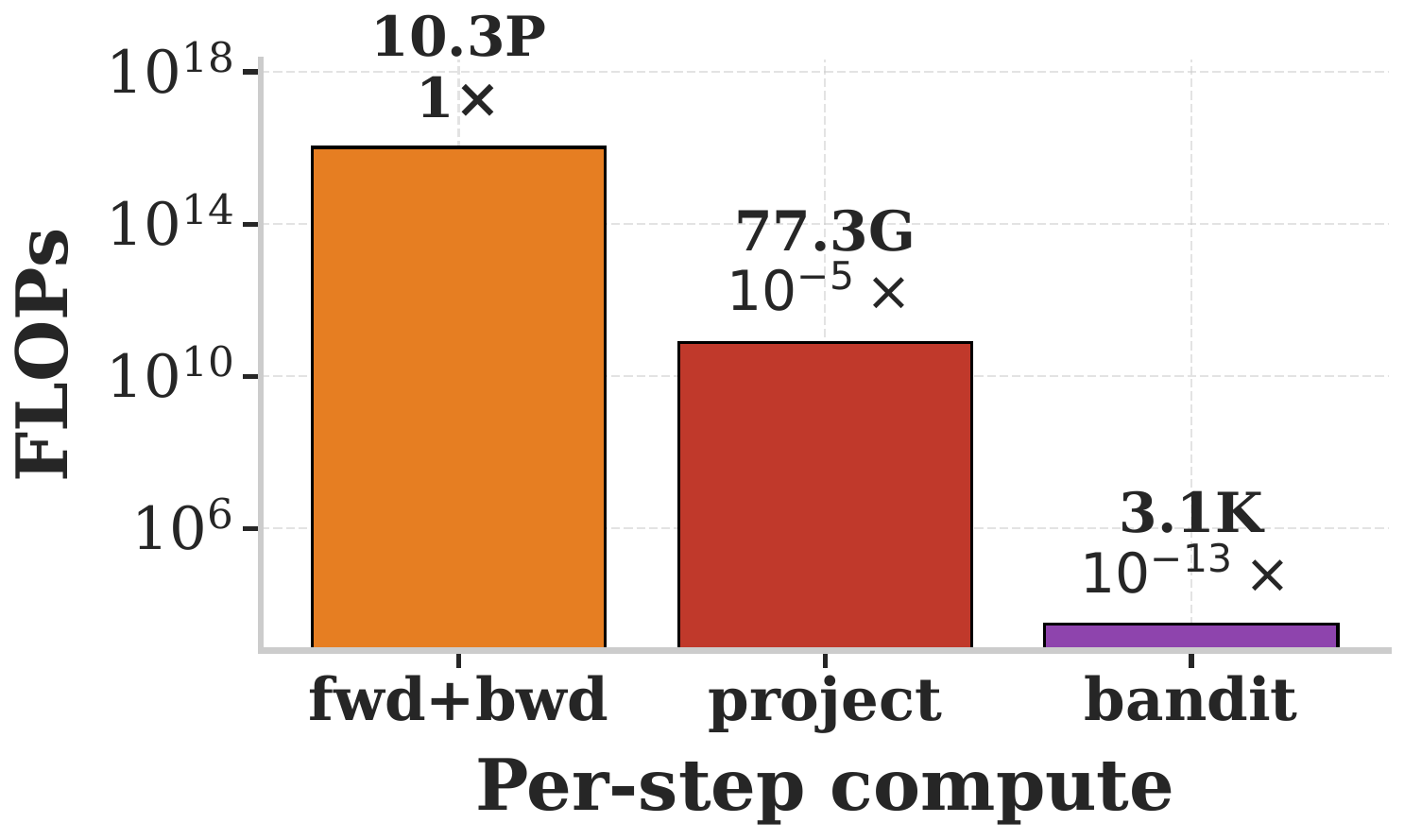}
    \caption{Per-step FLOPs.}
    \label{fig:walltime_flops}
\end{subfigure}
\caption{\textbf{Wall-clock overhead of \alg's transferability signal} (\texttt{Qwen3-1.7B-Base}, 4$\times$H100, batch 64 with $K{=}4$ rollouts). \textbf{(a)} Per-step decomposition into \emph{rollout} (vLLM generation), \emph{reward}, \emph{compute} (forward$+$backward$+$optimizer), \emph{other} (old-logprob$+$advantage), and \emph{transfer} (gradient projection $+$ bandit $Q$-update). \alg adds $1.06\,\mathrm{s}$ on top of a $115\,\mathrm{s}$ step ($0.9\%$); SEC has no \emph{transfer} slice. \textbf{(b)} Projection cost is essentially constant across the six training domains ($\sigma\!=\!2\,\mathrm{ms}$ on a $1.06\,\mathrm{s}$ mean), confirming that overhead depends only on parameter count and projection dimension, not on batch composition. \textbf{(c)} Analytic FLOP cost: projection is structurally ${\sim}\,10^{-5}\times$ the forward/backward compute and the bandit update ${\sim}\,10^{-13}\times$, well below the optimizer-step noise floor.}
\label{fig:walltime}
\vspace{-5pt}
\end{figure}

A central design goal of \alg is that the transferability signal should add negligible compute on top of standard GRPO. We verify this empirically by profiling each step on \texttt{Qwen3-1.7B-Base}, with a per-step batch of 64 prompts and $K{=}4$ rollouts on 4$\times$H100 GPUs. The profiled run uses $r{=}1024$; at our default $r{=}4096$ the projection cost grows by $4\times$ but remains far below the noise floor of the optimizer step, so the conclusions below are unchanged.

\alg's transferability machinery (gradient projection plus the bandit $Q$-update) adds only $1.06\,\mathrm{s}$ to a $115\,\mathrm{s}$ step---a $0.9\%$ overhead (Figure~\ref{fig:walltime_breakdown}). The dominant per-step costs remain rollout generation ($\sim$70\,s) and the forward/backward/optimizer pass ($\sim$35\,s); SEC, which uses no transferability signal, runs at the same wall-clock as \alg minus this $1.06\,\mathrm{s}$, confirming the bandit machinery itself is free and the entire cost lies in the projection.

The overhead is also stable across domains. Figure~\ref{fig:walltime_per_domain} reports projection wall-time per training step grouped by the domain of the sampled minibatch. The standard deviation across domains is $2\,\mathrm{ms}$ on a $1.06\,\mathrm{s}$ mean ($0.2\%$), confirming that the projection cost depends only on parameter count and projection dimension---not on batch composition or response length. The $\ell_2$ normalization in Eq.~\ref{eq:proj_grad} is a single vector-norm and contributes no measurable cost.

Figure~\ref{fig:walltime_flops} gives the analytic picture. The full forward/backward step costs roughly $10^{16}$ FLOPs ($\approx$10.3\,PFLOPs); the projection $P^\top g_t$ applied to the last $N{=}4$ transformer layers' gradients adds approximately $7.7\times 10^{10}$ FLOPs at the profiled $r{=}1024$ (a ${\sim}\,10^{-5}\times$ overhead), rising to ${\sim}\,3.1\times 10^{11}$ at our default $r{=}4096$, and the bandit $Q$-update is $\mathcal{O}(M)$ scalar operations across $M{=}6$ domains (${\sim}\,3\times 10^{3}$ FLOPs, ${\sim}\,10^{-13}\times$). Both are well below the noise floor of the optimizer step itself. The empirical $0.9\%$ wall-clock overhead measured in panel (a) is dominated by GPU--CPU transfer of the projected gradient and Python-side bookkeeping, not by the projection arithmetic itself, suggesting further reductions are possible but would have negligible practical effect.

Together, these measurements confirm \alg's design goal: the transferability signal is effectively free at training time. Wall-clock overhead is sub-1\%, stable across domains and batch compositions, and the analytic FLOP cost is $10^{-5}$ relative to the optimizer step itself. Practitioners can adopt \alg as a drop-in replacement for fixed-mixture or learnability-only sampling without modifying their training budget.

\subsection{A Pairwise View of Transferability}
\label{app:pairwise}
\begin{figure}[t]
\centering
\includegraphics[width=0.9\textwidth]{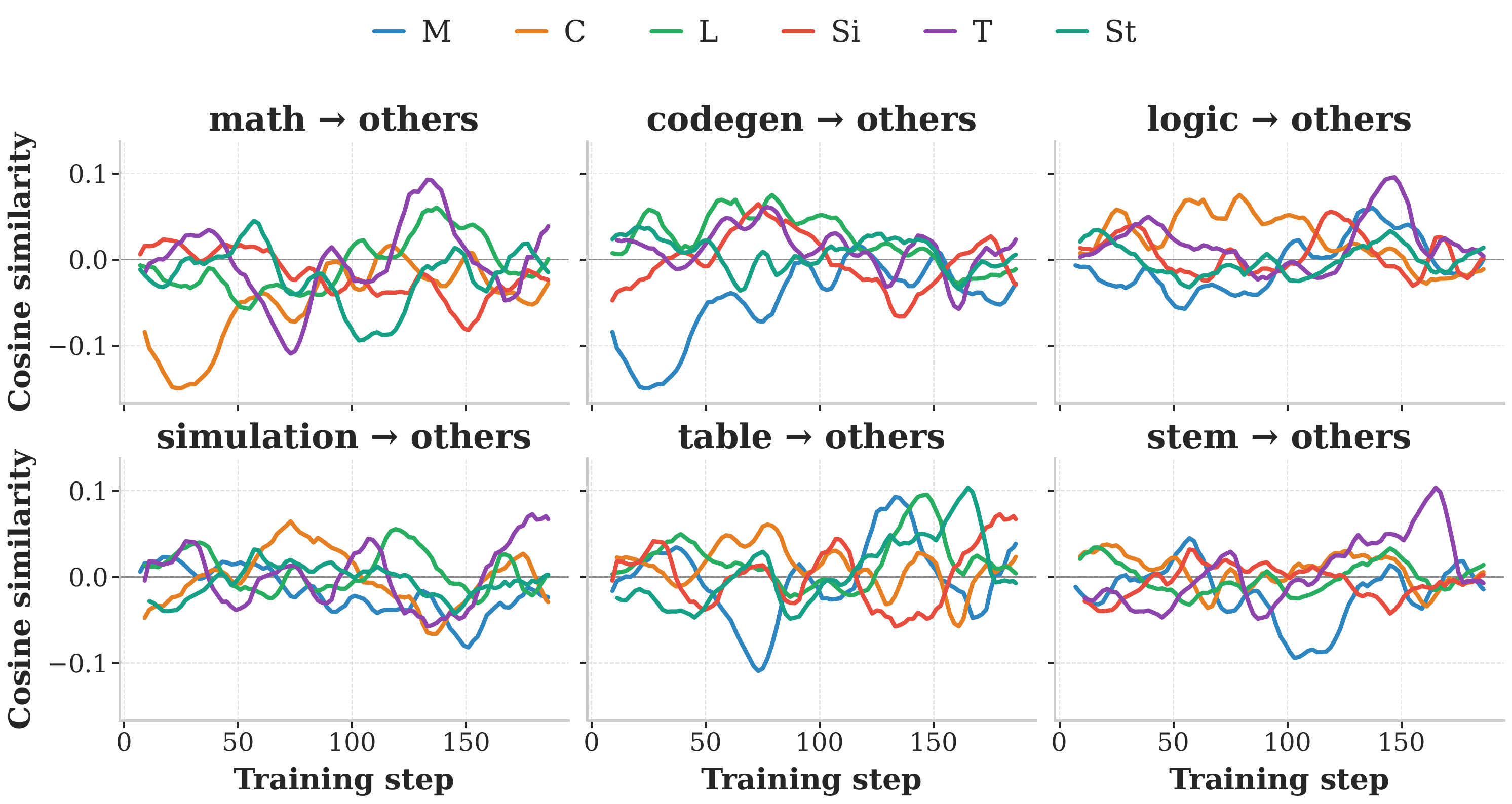}
\caption{\textbf{Pairwise gradient cosine similarity over training, per source domain.} Each panel plots the cosine similarity between the projected-gradient EMA $\mathbf{h}_m^{(t)}$ of one \emph{source} domain $m$ (panel title) and the EMAs of the five other \emph{target} domains (line color), averaged across three \alg{} seeds and smoothed with a 10-step running mean. The aggregate transferability signal $T_m^{(t)}$ consumed by the bandit (Eq.~\ref{eq:minmax_transfer}) is a cross-domain--normalized transform of each row's mean cosine (Eq.~\ref{eq:raw_transfer}). Absolute magnitudes are small ($|\cos|\!\lesssim\!0.15$)---domain pairs are only weakly aligned in the projected subspace---but the \emph{relative} structure is stable and coherent: \emph{table} acts as a hub, ending positively aligned with \emph{simulation} and \emph{stem} (its two strongest bonds) and with \emph{logic}, which is why it carries the highest aggregate $T$ in Figure~\ref{fig:ttilde_per_domain}; \emph{math} and \emph{codegen}, by contrast, sit at negative cosine with nearly every domain---and most negatively with \emph{each other}---explaining their lowest aggregate $T$.}
\label{fig:pairwise_transfer}
\end{figure}
Figure~\ref{fig:pairwise_transfer} decomposes the aggregate $T_m^{(t)}$ signal into its underlying per-pair cosines, with one panel per source domain. Two observations are worth emphasizing. First, the absolute magnitudes are small: nearly all pairs stay within $|\cos|\!\lesssim\!0.15$ across training, confirming that gradient agreement in the projected subspace is weak in absolute terms. The cross-domain normalization of Eq.~(\ref{eq:minmax_transfer}) is therefore essential---raw cosines alone would be uninformative as a curriculum signal, but the \emph{relative} ordering across domains is robust enough to drive sample reallocation. Second, that ordering has a clear geometric structure rather than being noise. \emph{Table} behaves as a hub: its gradients end positively aligned with \emph{simulation} and \emph{stem} (its two largest cosines) and with \emph{logic}, so training on table moves the policy in directions that also help those domains---the geometric origin of table's top-ranked $T$ in Figure~\ref{fig:ttilde_per_domain}. \emph{Math} and \emph{codegen}, by contrast, sit at negative cosine with almost every other domain, and their \emph{mutual} cosine is the most negative pair in the matrix; their updates pull the policy in idiosyncratic directions that do not benefit---and often oppose---the rest of the mixture, which is why they carry the lowest aggregate $T$ and \alg{} samples them least. We read this isolation as a pretraining effect: base models, and \texttt{Qwen3} in particular, are already saturated on math and code, so RL there sharpens an already-formed skill along directions roughly orthogonal to the shared reasoning gradients the other domains move along.

\subsection{Scaling Experiments}
\label{app:model_scaling}
\begin{table*}[t]
\caption{Additional results across two model sizes (\texttt{Qwen3-0.6B-Base} and \texttt{Qwen3-4B-Base}) using the same training data setup and baselines as the main table. Each configuration is run with a single seed.}
\label{tab:model_size_results}
\centering
\small
\setlength{\tabcolsep}{3pt}
\renewcommand{\arraystretch}{1.10}
\resizebox{\textwidth}{!}{
\begin{tabular}{lccccccccccccccc}
\toprule
\textbf{Method} & \multicolumn{2}{c}{\textbf{Codegen}} & \multicolumn{2}{c}{\textbf{Logic}} & \multicolumn{2}{c}{\textbf{Math}} & \multicolumn{3}{c}{\textbf{Simulation}} & \multicolumn{2}{c}{\textbf{STEM}} & \multicolumn{3}{c}{\textbf{Table}} & \textbf{All} \\
\cmidrule(lr){2-3} \cmidrule(lr){4-5} \cmidrule(lr){6-7} \cmidrule(lr){8-10} \cmidrule(lr){11-12} \cmidrule(lr){13-15} \cmidrule(lr){16-16}
 & HE & MBPP & ARC-AGI & Zebra & MATH & AIME & CodeI/O & CruxEval-I & CruxEval-O & GPQA & SuperGPQA & MH & FinQA & HiTab & \textit{macro avg.} \\
\midrule
\multicolumn{16}{l}{\textit{\textbf{Qwen3-0.6B}}} \\
Random & $\mathbf{32.9}$ & $\mathbf{37.0}$ & $0.3$ & $5.6$ & $28.6$ & $1.5$ & $\mathbf{5.1}$ & $26.9$ & $28.6$ & $27.7$ & $15.6$ & $4.2$ & $0.4$ & $36.5$ & $18.1$ \\
SEC & $30.5$ & $32.4$ & $0.3$ & $2.6$ & $44.4$ & $0.9$ & $2.8$ & $\mathbf{33.4}$ & $35.2$ & $\mathbf{29.5}$ & $15.0$ & $12.0$ & $3.8$ & $37.9$ & $19.9$ \\
\cellcolor{blue!10}\textbf{\alg} & \cellcolor{blue!10}$32.0$ & \cellcolor{blue!10}$35.6$ & \cellcolor{blue!10}$\mathbf{0.5}$ & \cellcolor{blue!10}$\mathbf{13.4}$ & \cellcolor{blue!10}$\mathbf{47.0}$ & \cellcolor{blue!10}$\mathbf{1.7}$ & \cellcolor{blue!10}$2.6$ & \cellcolor{blue!10}$29.6$ & \cellcolor{blue!10}$\mathbf{35.6}$ & \cellcolor{blue!10}$27.5$ & \cellcolor{blue!10}$\mathbf{16.9}$ & \cellcolor{blue!10}$\mathbf{12.3}$ & \cellcolor{blue!10}$\mathbf{4.5}$ & \cellcolor{blue!10}$\mathbf{41.5}$ & \cellcolor{blue!10}$\mathbf{21.6}$ \\
\midrule
\multicolumn{16}{l}{\textit{\textbf{Qwen3-4B}}} \\
Random & $83.2$ & $\mathbf{67.4}$ & $\mathbf{4.5}$ & $36.4$ & $74.4$ & $10.8$ & $4.2$ & $59.6$ & $47.6$ & $41.3$ & $30.5$ & $44.9$ & $37.0$ & $61.4$ & $43.2$ \\
SEC & $81.1$ & $66.6$ & $4.2$ & $35.6$ & $73.4$ & $\mathbf{12.9}$ & $5.2$ & $59.5$ & $\mathbf{62.7}$ & $36.7$ & $29.9$ & $45.4$ & $38.9$ & $65.4$ & $43.8$ \\
\cellcolor{blue!10}\textbf{\alg} & \cellcolor{blue!10}$\mathbf{83.5}$ & \cellcolor{blue!10}$\mathbf{67.4}$ & \cellcolor{blue!10}$2.6$ & \cellcolor{blue!10}$\mathbf{36.9}$ & \cellcolor{blue!10}$\mathbf{76.2}$ & \cellcolor{blue!10}$12.3$ & \cellcolor{blue!10}$\mathbf{5.6}$ & \cellcolor{blue!10}$\mathbf{61.4}$ & \cellcolor{blue!10}$62.1$ & \cellcolor{blue!10}$\mathbf{42.5}$ & \cellcolor{blue!10}$\mathbf{32.0}$ & \cellcolor{blue!10}$\mathbf{47.7}$ & \cellcolor{blue!10}$\mathbf{39.7}$ & \cellcolor{blue!10}$\mathbf{70.1}$ & \cellcolor{blue!10}$\mathbf{45.4}$ \\
\bottomrule
\end{tabular}
}
\end{table*}

Table~\ref{tab:model_size_results} shows that \alg's gains generalize across model scales within the Qwen3 family. On \texttt{Qwen3-0.6B-Base}, \alg reaches $21.6$ macro accuracy against $18.1$ (Random) and $19.9$ (SEC), gaps of $+3.5$ and $+1.7$, ranking first on $9/14$ benchmarks. On \texttt{Qwen3-4B-Base}, \alg reaches $45.4$ versus $43.2$ (Random) and $43.8$ (SEC), gaps of $+2.2$ and $+1.6$, ranking first on $10/14$. Combined with the $+2.1$/$+1.8$ over Random/SEC at 1.7B in Table~\ref{tab:main_results}, the per-scale gains over SEC are comparable ($+1.6$ to $+1.8$) and \alg improves over Random at every scale ($+2.1$ to $+3.5$): neither vanishing at the smaller backbone where rollouts are noisier, nor washing out at the larger one where the gradient geometry is higher-dimensional. We attribute this scale-invariance to the signal being computed from each model's own training gradients rather than any fixed external probe---as the policy changes shape with scale, $T$ adapts with it.

\section{More Examples}
\label{app:more_examples}

The following examples compare rollouts from models trained with \alg{} and with the \textit{Random} curriculum baseline. We sample one example each from \textsc{MATH-500}~\citep{hendrycks2021measuring}, \textsc{GPQA-Diamond}~\citep{rein2024gpqa}, and \textsc{Zebra-Puzzle}~\citep{lin2025zebralogic}, drawn from final checkpoints of \texttt{Qwen3-1.7B-Base} runs. The correct final answer is highlighted in \textcolor{green!60!black}{green} and an incorrect one in \textcolor{red}{red}. These examples are illustrative rather than systematic: they showcase the kinds of qualitative differences we observe between the two curricula on representative problems from three of the six training domains, with \alg{}-trained models more consistently following through the right chain of reasoning where Random-trained models stall on early misidentifications or fail to invoke the relevant principle (relativistic time dilation in the GPQA case, parity propagation in the Zebra case).

\begin{tcolorbox}[enhanced, breakable, title={Example 1: Algebra (\texttt{MATH}, \texttt{Qwen3-1.7B-Base}) \\ Random vs. \alg}, colframe=black!70, colback=gray!5]
\textbf{Question.} What is the smallest positive integer $n$ such that all the roots of $z^4 + z^2 + 1 = 0$ are $n^{\text{th}}$ roots of unity? \\[0.5em]
\textbf{Ground-truth answer.} $\boxed{6}$ \\[0.5em]
\textbf{Response from Random-trained model.} Substituting $w = z^2$ gives $w^2 + w + 1 = 0$, so:
$$ w = \frac{-1 \pm i\sqrt{3}}{2} $$
$$ \cdots $$
\text{The model identifies } $\frac{-1 + i\sqrt{3}}{2} = e^{i\pi/3}$ \text{ (incorrect: the argument is } $2\pi/3$\text{).}
$$ \cdots $$
Taking square roots under this misidentification yields:
$$ z \in \{e^{\pm i\pi/6},\, e^{\pm i 5\pi/6}\} $$
\text{The model concludes these are 12th roots of unity.}
\subsubsection*{Final Answer}
$$ \boxed{\textcolor{red}{12}} $$ \\[0.5em]
\textbf{Response from \alg-trained model.} Substituting $w = z^2$ gives $w^2 + w + 1 = 0$, so:
$$ w = \frac{-1 \pm i\sqrt{3}}{2} = e^{\pm 2\pi i/3} $$
$$ \cdots $$
The four roots of the original quartic are:
$$ z = \pm e^{\pm i\pi/3} $$
$$ \cdots $$
Since $e^{\pm i\pi/3}$ are primitive 6th roots of unity, and $-e^{\pm i\pi/3}$ are also 6th roots of unity, all four roots are 6th roots of unity.
\subsubsection*{Final Answer}
$$ \boxed{\textcolor{green!60!black}{6}} $$ \\[0.5em]
\end{tcolorbox}

\begin{tcolorbox}[enhanced, breakable, title={Example 2: Particle Physics (\texttt{GPQA-Diamond}, \texttt{Qwen3-1.7B-Base}) \\ Random vs. \alg}, colframe=black!70, colback=gray!5]
\textbf{Question.} In the CERN Bubble Chamber a decay $X^{0} \to Y^{+} Z^{-}$ occurs with proper lifetime $\tau_0 = 8 \times 10^{-16}$\,s. What minimum spatial resolution is needed to observe at least 30\% of the decays, given an energy of $27$\,GeV and mass $m_{X^0} = 3.41$\,GeV?

\quad A) $2.08 \times 10^{-3}$\,m \quad B) $2.08 \times 10^{-6}$\,m \quad C) $2.08 \times 10^{-1}$\,m \quad D) $2.08 \times 10^{-9}$\,m \\[0.5em]
\textbf{Ground-truth answer.} $\boxed{\mathrm{B}}$ \\[0.5em]
\textbf{Response from Random-trained model.} The model attempts to set up a decay-rate calculation:
$$ \lambda = \frac{1}{\tau_0} $$
$$ \cdots $$
\text{It tries to convert ``number of decays'' into a spatial resolution via } $\Delta x = v \cdot t$, \text{ leaving the decay count } $N$ \text{ as a free parameter.}
$$ \cdots $$
\text{The chain stalls without setting up the relativistic boost. Across four rollouts,} \\
\text{the model selects A or D three times.}
\subsubsection*{Final Answer}
$$ \boxed{\textcolor{red}{\mathrm{D}}} $$ \\[0.5em]
\textbf{Response from \alg-trained model.} Recognizing this as a time-dilation problem, we compute the Lorentz factor:
$$ \gamma = \frac{E}{m} = \frac{27}{3.41} \approx 7.92 $$
$$ \cdots $$
The lab-frame decay length is:
$$ d = \gamma\, \tau_0\, c \approx 7.92 \cdot 8 \times 10^{-16}\,\mathrm{s} \cdot 3 \times 10^{8}\,\mathrm{m/s} $$
$$ d \approx 1.9 \times 10^{-6}\,\mathrm{m} $$
$$ \cdots $$
This matches option B to within rounding.
\subsubsection*{Final Answer}
$$ \boxed{\textcolor{green!60!black}{\mathrm{B}}} $$ \\[0.5em]
\end{tcolorbox}

\begin{tcolorbox}[enhanced, breakable, title={Example 3: Constraint Satisfaction (\texttt{Zebra-Puzzle}, \texttt{Qwen3-1.7B-Base}) \\ Random vs. \alg}, colframe=black!70, colback=gray!5]
\textbf{Question (abridged).} Two positions, each with attributes Beverage $\in$ \{iced-tea, tea\}, Hobby $\in$ \{traveling, writing\}, Movie-Genre $\in$ \{fantasy, spy\}, Music-Genre $\in$ \{d\&b, hip-hop\}, Nationality $\in$ \{australian, russian\}, Pet $\in$ \{ferret, fish\}.

\textit{Clues:} \texttt{Beverage:tea} is on the far right; \texttt{Hobby:traveling} and \texttt{Music-Genre:hip-hop} have the same parity; \texttt{Music-Genre:d\&b} is left of \texttt{Hobby:traveling}; \texttt{Movie-Genre:fantasy} is left of \texttt{Movie-Genre:spy}; \texttt{Music-Genre:hip-hop} and \texttt{Nationality:australian} have the same parity; \texttt{Beverage:iced-tea} is left or right of \texttt{Pet:ferret}. \\[0.5em]
\textbf{Ground-truth grid.} \\
\texttt{[[1, iced-tea, writing, fantasy, d\&b, russian, fish],}\\
\texttt{\phantom{[}[2, tea, traveling, spy, hip-hop, australian, ferret]]} \\[0.5em]
\textbf{Response from Random-trained model.} The model lists and restates the clues but never propagates the parity constraints.
$$ \cdots $$
\text{One rollout overwrites the schema with the example header from the prompt}\\
\text{(``Nationality, Job''); another invents extra rows.}
$$ \cdots $$
Average cell-level accuracy across four rollouts: $\sim$36\%.
\subsubsection*{Final Answer}
\textcolor{red}{Inconsistent grid (schema mismatch).} \\[0.5em]
\textbf{Response from \alg-trained model.} All four rollouts return the requested JSON in the correct schema:

{\footnotesize
\texttt{<answer>\{"header": ["Position", "Beverage", "Hobby",}\\
\texttt{"Movie-Genre", "Music-Genre", "Nationality", "Pet"],}\\
\texttt{"rows": [["1", "iced-tea", "writing", "fantasy", "d\&b",}\\
\texttt{"russian", "fish"], ["2", "tea", "traveling", "spy",}\\
\texttt{"hip-hop", "australian", "ferret"]]\}</answer>}}

\subsubsection*{Final Answer}
\textcolor{green!60!black}{Correct grid (89--100\% cell-level accuracy across rollouts).} \\[0.5em]
\end{tcolorbox}

\end{document}